\documentclass[11pt]{article}

\newcommand{\affilsup}[1]{\textsuperscript{#1}}
\usepackage{authblk}

\usepackage[final]{acl}

\usepackage{times}
\usepackage{latexsym}
\usepackage[T1]{fontenc}
\usepackage{amsmath,amssymb,mathtools}
\usepackage{graphicx}
\usepackage{xcolor}
\usepackage{xspace}

\usepackage[most]{tcolorbox}

\usepackage{dblfloatfix} 
\usepackage{float}       
\usepackage{algorithm}
\usepackage{algpseudocode}

\usepackage{booktabs}
\usepackage{multirow}
\usepackage{multicol}
\usepackage{array}
\usepackage{makecell}
\usepackage{tabularx}
\usepackage{colortbl}
\usepackage{threeparttable}
\usepackage{adjustbox}

\usepackage{caption}
\usepackage{subcaption}

\usepackage{enumitem}
\setlist[itemize]{leftmargin=1.2em,topsep=2pt,itemsep=1.5pt,parsep=0pt}

\usepackage{titletoc}

\definecolor{headerblue}{HTML}{6495ED}
\definecolor{headerbg}{HTML}{B0C4DE}
\definecolor{boxbg}{HTML}{FDF6E3}
\definecolor{correct}{HTML}{2E7D32}
\definecolor{wrong}{HTML}{C62828}
\definecolor{gtcolor}{HTML}{B8860B}
\definecolor{stepblue}{HTML}{1565C0}
\definecolor{commitpurple}{HTML}{6A1B9A}
\definecolor{reasongray}{HTML}{555555}
\definecolor{divgray}{HTML}{CCCCCC}
 
\newcommand{\method}{ActiveMedAgent\xspace}

\newcommand{\best}[1]{\textbf{#1}}
\newcommand{\improve}[1]{\textcolor{correct}{#1}}
\newcommand{\degrade}[1]{\textcolor{wrong}{#1}}

\newcommand{\agentres}[1]{\textcolor{correct}{\textbf{Agent:}} #1}
\newcommand{\oracleres}[1]{\textcolor{wrong}{\textbf{Full Modality:}} #1}
\newcommand{\step}[1]{\textcolor{stepblue}{\textbf{#1}}}
\newcommand{\commitlabel}{\textcolor{commitpurple}{\textbf{Commit:}}}
\newcommand{\why}[1]{\textcolor{reasongray}{\textit{#1}}}

\newtcolorbox{hcasebox}[1]{%
  enhanced,
  title={#1},
  fonttitle=\bfseries\small,
  coltitle=black,
  colbacktitle=headerbg,
  colback=boxbg,
  colframe=headerblue,
  boxrule=0.8pt,
  arc=2pt,
  left=5pt, right=5pt, top=4pt, bottom=4pt,
  before skip=6pt, after skip=6pt,
  fontupper=\small,
  toptitle=2pt, bottomtitle=2pt,
}

\newtcolorbox{casebox}[1]{%
  enhanced,
  breakable,
  width=\linewidth,
  title={#1},
  fonttitle=\bfseries\small,
  coltitle=black,
  colbacktitle=headerbg,
  colback=boxbg,
  colframe=headerblue,
  boxrule=0.8pt,
  arc=2pt,
  left=5pt, right=5pt, top=4pt, bottom=4pt,
  before skip=6pt, after skip=6pt,
  fontupper=\small,
  toptitle=2pt, bottomtitle=2pt,
}

\usepackage[T1]{fontenc}

\usepackage[utf8]{inputenc}

\usepackage{microtype}

\usepackage{inconsolata}

\usepackage{graphicx}

\title{\method: Cost-Aware Trajectory Learning \\for Multimodal Medical Diagnosis}

\author{Weiwei Ma\affilsup{1} \quad Xiaobing Yu\affilsup{4}$^{\dag}$ \quad Peijie Qiu\affilsup{1}\thanks{This work was done prior to joining Amazon.} \quad Jin Yang\affilsup{2}\\
  Zhaoqi An\affilsup{1} \quad Xuanzhao Dong\affilsup{3} \quad Xiaoqi Zhao\affilsup{4}$^{\dag}$ \quad Xiaofeng Liu\affilsup{4}$^{\dag}$\\
  $^1$\large{Washington University in St. Louis};  \quad $^2$ \large{Icahn School of Medicine at Mount Sinai}; \\ \quad $^3$\large{Arizona State University}; \quad $^4$\large{Yale University}\\
  \texttt{m.weiwei@wustl.edu} \\
  \texttt{\{xiaobing.yu, xiaoqi.zhao, xiaofeng.liu\}@yale.edu}\\
$^{\dag}$ co-corresponding author\\
\textbf{Code}: \url{https://github.com/VisualReasoner/activeagent}\\
}

\begin{document}
\maketitle
\begin{abstract}
Clinical diagnosis is inherently sequential: clinicians
escalate from cheap to costly tests only when additional
evidence is expected to resolve diagnostic uncertainty.
We present \method, a framework that brings this
cost-aware sequential logic to multimodal medical AI.
Given a frozen, API-accessed vision-language model, \method
tracks probability distributions over candidate diagnoses and scores
each acquisition by its per-step diagnostic utility minus cost.
A lightweight MLP controller is then trained offline
on these scored trajectories, learning when to request
additional evidence and when to commit.
Across three commonly used benchmarks, trajectory-based policy
learning consistently outperforms both unguided acquisition
and full-modality baselines. Notably, we identify an
information overload effect. In 175 cases, the agent produces
a correct diagnosis with fewer channels while the
full-modality baseline fails, showing that learning
\emph{what to omit} can be as important as learning what
to acquire.
\end{abstract}

\begin{figure*}[t]
\centering
\begin{hcasebox}{Representative Successful Case: OLIVES / GPT-4o-mini}
\begin{minipage}[t]{0.26\textwidth}
\textbf{Ground truth}\\[3pt]
Atrophy, Ir\,Hrf, Partial\,Vitreous
\end{minipage}\hfill
\begin{minipage}[t]{0.42\textwidth}
\textbf{Acquisition trace}\\[3pt]
\step{Step 0:} +clinical\_measurements (\$20) leaves a diffuse differential but narrows the case toward vitreous and atrophy patterns. \newline
\step{Step 1:} +biomarker\_hints (\$100) moves \textbf{Atrophy, Ir\,Hrf, Partial\,Vitreous} to rank 1. \newline
\commitlabel\ The policy stops because OCT and extra slices have low expected value after the biomarker update.
\end{minipage}\hfill
\begin{minipage}[t]{0.26\textwidth}
\textbf{Outcome}\\[3pt]
\agentres{2 channels, MRR=1.0, correct, \$120.} \newline
\oracleres{4 channels, MRR=25\%, wrong, \$570.} \newline
\why{Stopping avoids extra image evidence that distracts the model.}
\end{minipage}
\end{hcasebox}
\caption{\textbf{Successful selective acquisition case.}
PolicyNet commits after two low-cost evidence channels, while Full Modality receives all four channels and ranks the correct diagnosis lower. This illustrates the information-overload mechanism that motivates cost-aware trajectory learning.}
\label{fig:successful_case}
\end{figure*}

\section{Introduction}

We start from a counterintuitive failure mode: in 175 test cases, a frozen VLM is correct with selected evidence but wrong with all evidence, suggesting that multimodal diagnosis is also an evidence-control problem.
Medical multimodal reasoning is commonly evaluated as a static prediction problem. A model receives an image, sometimes paired with supporting text, and produces a diagnosis after all evidence has already been revealed. Recent medical vision-language systems such as Med-Flamingo~\cite{moor2023med} and LLaVA-Med~\cite{li2023llava} demonstrate strong performance under this paradigm, but they are typically assessed in settings where the visual modality is always present, and the interaction reduces to a form of medical visual question-answering (VQA). This abstraction is convenient, yet it omits a defining property of real diagnosis: evidence is acquired progressively through selective escalation.

In clinical practice, physicians rarely begin with every modality at once. They first inspect low-cost evidence such as demographics, presenting complaint, medical history, and basic examination findings. Imaging and other expensive investigations are ordered only when earlier evidence leaves material uncertainty. A model that reaches the correct answer only after consuming every modality is therefore not necessarily acting intelligently. In many cases, it is simply overusing information. The clinically meaningful question is thus economic and predictive: what is the minimum amount of evidence needed to arrive at the correct diagnosis? This observation turns multimodal diagnosis into a sequential reasoning problem. At each step, the diagnostic system must reason about the currently available evidence, maintain a belief distribution over candidate diagnoses, and decide whether acquiring an additional modality is expected to improve the diagnosis sufficiently to justify its cost. The objective is no longer only to predict correctly after all evidence is available, but to decide when additional evidence is necessary.

We study this setting as cost-aware multimodal diagnostic escalation. The central idea of this work is to learn from \emph{diagnostic belief trajectories}. As new evidence is acquired, the model produces a sequence of belief states over diagnoses. These trajectories reveal how each modality affects the differential: some acquisitions resolve ambiguity, some are redundant, and some introduce distraction or information overload. In contrast to static multimodal learning, which observes only the final input and label, trajectory learning supervises how reasoning evolves over time. This enables the system to improve not only prediction accuracy, but also the decision process of whether more evidence should be acquired.

Based on this formulation, we present \method, a framework for cost-aware sequential multimodal diagnosis. The model starts from presentation-time evidence and iteratively chooses between acquiring one additional information channel or committing to a ranked differential diagnosis. A lightweight controller is learned from previously collected trajectories and operates over compact diagnostic states. Importantly, the backbone vision-language model remains unchanged. The improvement arises from learning better evidence acquisition strategies rather than modifying the underlying diagnostic model.

We evaluate \method on NEJM~\cite{fajtl2024trustworthy}, MIDAS~\cite{chiou2025multimodal}, and OLIVES~\cite{prabhushankar2022olives} under clinically ordered evidence hierarchies. The results reveal that active acquisition improves over passive diagnosis, but naive acquisition is often brittle, and that the benefit of learning depends strongly on model capacity and domain structure.
Figure~\ref{fig:successful_case} gives the paper's core example: the learned policy stops after two low-cost OLIVES channels, reaches the correct diagnosis, and avoids later OCT evidence that causes the full-modality baseline to fail.

\noindent In summary, our contributions are fourfold:
\textit{(i)} we introduce a structured, hierarchy-constrained formulation of multimodal diagnosis that replaces free-form query generation with a finite, clinically grounded action space;
\textit{(ii)} we propose trajectory-level supervision over belief state transitions, enabling learning of acquisition policies from intermediate reasoning signals rather than final predictions alone;
\textit{(iii)} we demonstrate that learning acquisition policies offline yields substantially more reliable decisions than prompt-based or self-reflective strategies, for example, improving NEJM accuracy from 18.6\% to 31.2\% on GPT-4o-mini, where naive acquisition fails; and
\textit{(iv)} we identify and quantify an information overload effect, with 175 cases where selective acquisition is correct while the full-information baseline is wrong, showing that additional modalities can degrade reasoning and motivating evaluation based on accuracy and cost rather than accuracy alone.

\section{Related Work}
\textbf{Sequential, cost-aware diagnosis.} Prior medical VLM and VQA benchmarks typically assume all modalities are available upfront~\cite{moor2023med,li2023llava,lin2023medical,ma2026u}. Closer to our setting, recent work studies interactive/sequential diagnosis and incorporates test costs~\cite{li2024mediq,nori2025sequential}. In particular, SDBench transforms NEJM clinicopathological conference cases into stepwise diagnostic encounters under explicit cost constraints, and MAI-DxO provides a model-agnostic orchestration layer (e.g., simulating a physician panel) to decide what information to request and when to commit~\cite{nori2025sequential}. These systems are strong baselines for sequential decision-making, but (i) primarily operate in a free-form question/test-request space that is harder to audit clinically, and (ii) do not target structured acquisition over a fixed, clinically meaningful hierarchy of evidence channels. \method focuses on this structured setting, enabling transparent policies and controlled cost-accuracy evaluation in multimodal diagnosis.

\noindent\textbf{Clinical LLM evaluation and decision support.} Recent work evaluates LLMs and VLMs on medical exams and clinical reasoning, and highlights both promise and reliability gaps~\cite{Kung2022PerformanceOCA,Singhal2022LargeLMA,Singhal2025TowardEMA,Nori2023CapabilitiesOGA,mcduff2025towards,wan2024humans,griot2025large,mccoy2025language,thapa2025disentangling,griot2025pattern,chen2024clinicalbench,hao2025medpair,jeong2024medical}. Our setting also connects to earlier diagnostic decision support and cost-aware clinical workflow integration~\cite{barnett1987dxplain,berner1994performance,elhanan1996integrating,detmer1997using,barnett1998dxplain,bauer2002internal,elkin2010introduction,nee2010clinical}.

\noindent\textbf{Active acquisition beyond explicit tool use.} While tool-use frameworks motivate interleaving reasoning with actions~\cite{yao2022react,schick2023toolformer}, our controller does not rely on general-purpose external tools. Instead, it learns when to escalate to additional modalities under explicit costs, connecting to active feature acquisition under cost constraints~\cite{li2021active,yang2025active,yu2025fm}.

\noindent\textbf{Foundation models, data, and probabilistic beliefs.} Our work builds on modern general-purpose and open foundation models and their training/evaluation ecosystems~\cite{achiam2023gpt,hurst2024gpt,touvron2023llama,touvron2023llama2,groeneveld-etal-2024-olmo,soldaini-etal-2024-dolma,weber2024redpajama,Liu2024InfiniGram,hendrycks2020measuring,srivastava2023beyond,yang2018hotpotqa}. Finally, extracting actionable beliefs from natural-language outputs relates to work on probabilistic reasoning and uncertainty in language models~\cite{paruchuri2024odds,feng2024bird,nafar2025extracting}. Unlike prior work, our approach separates reasoning (backbone VLM) from decision-making (learned policy).

\section{Method}
\label{sec:method}

\begin{figure*}[t]
  \centering
  \includegraphics[width=0.92\textwidth]{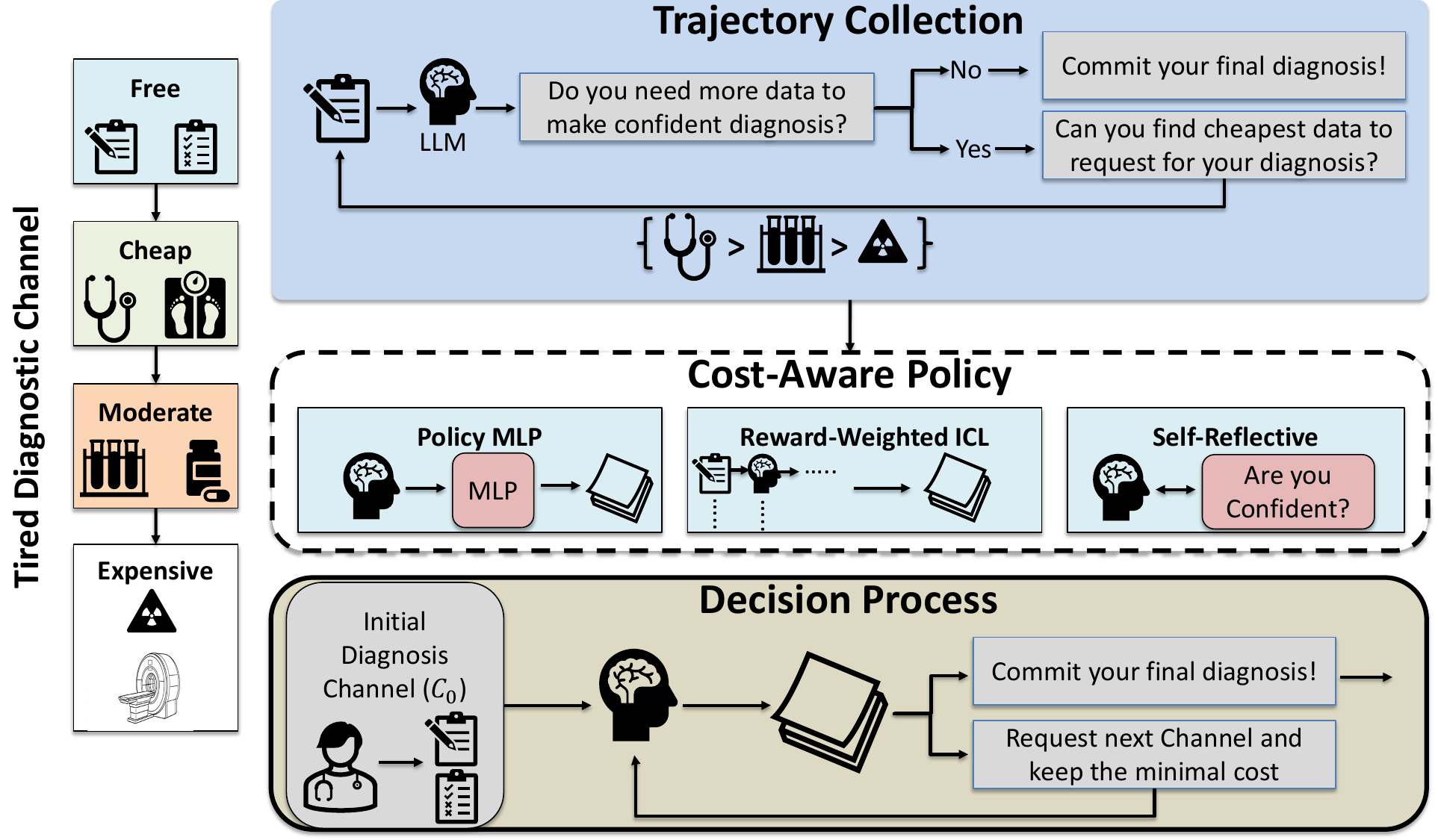}
  \caption{\textbf{\method framework overview.}
  \textit{Left:} Diagnostic channels organized into a tiered cost hierarchy.
  \textit{Top:} Trajectory collection via structured VLM interaction.
  \textit{Middle:} Cost-aware policy learning for channel acquisition.
  \textit{Bottom:} Inference-time decision loop with adaptive stopping.}
  \label{fig:system_overview}
\end{figure*}

Multimodal medical AI usually assumes simultaneous access to all evidence.
Clinical diagnosis is sequential: each test has cost, delay, and potential
to distract the model when ordered unnecessarily.

Existing approaches to this problem either require environment simulators
or access to model internals unavailable in API-served foundation
models~\cite{yu2023deep,shim2018joint}, treat modality selection as a
static feature-selection problem that ignores how beliefs evolve as
evidence accumulates~\cite{li2021active}, or encode acquisition logic entirely
in natural-language prompting with no learned, auditable decision
boundary~\cite{nori2025sequential}.

\method studies structured evidence acquisition over a fixed clinical
hierarchy. As evidence is revealed, the frozen VLM produces belief states;
each transition is scored by diagnostic improvement minus acquisition cost
(Eq.~\ref{eq:utility}). A lightweight MLP controller (PolicyNet) is trained
offline on these scored trajectories and uses only compact state features
(channel mask, belief entropy, confidence gap, cumulative cost).

This design yields an acquisition policy that is \emph{decoupled} from the
backbone (retrained or replaced without modifying the VLM),
\emph{auditable} (a small MLP over interpretable features), and
\emph{trajectory-supervised} (able to distinguish acquisitions that resolve
ambiguity from those that are redundant or harmful---a signal unavailable
to prompt-only systems).  We refer to
Figure~\ref{fig:system_overview} for an overview of the full pipeline and
Algorithm~\ref{alg:workflow} for the complete acquisition loop.

\subsection{Problem Formulation}

A clinical case $x = (\{v_c\}_{c \in \mathcal{C}},\; y)$ consists of diagnostic channels $\mathcal{C} = \mathcal{C}_0 \cup \mathcal{C}_R$ and a ground-truth diagnosis $y \in \mathcal{Y}$, where $|\mathcal{Y}| = K$.
Free channels $\mathcal{C}_0$ are available at presentation; requestable channels $\mathcal{C}_R$ each have acquisition cost $\kappa(c) \geq 0$ and modality type $\tau(c) \in \{\texttt{text}, \texttt{image}\}$.
At step $t$, observed evidence is $E_t = \{v_c : c \in \mathcal{C}_0 \cup \mathcal{A}_t\}$, where $\mathcal{A}_t \subseteq \mathcal{C}_R$ records acquired channels.
The VLM backbone maps evidence to a belief distribution $\mathbf{b}_t = f_\phi(E_t) \in \Delta^{K-1}$ over candidates, and the agent selects
$a_t \in (\mathcal{C}_R \setminus \mathcal{A}_t) \cup \{COMMIT\}$.
We maximize the quality-cost objective
\begin{equation}
\begin{aligned}
  J(\pi) = \mathbb{E}\!\Big[&
  \mathrm{RR}(\mathbf{b}_T, y) \\
  &- \lambda \sum_{t < T} \hat{\kappa}(a_t)\,
  \mathbf{1}[a_t \neq COMMIT]\Big],
\end{aligned}
\label{eq:objective}
\end{equation}
where $\mathrm{RR}(\mathbf{b}_T, y) = 1/\mathrm{rank}_{\mathbf{b}_T}(y)$ is reciprocal rank, $\hat{\kappa}(a_t) = \kappa(a_t)/\max_c \kappa(c)$ is normalized cost, and $\lambda$ controls the trade-off.
We use reciprocal rank because ranked differentials are top-heavy: moving the correct diagnosis from rank 5 to rank 2 should matter, but less than moving it to rank 1, and RR is label-set-size independent compared with NDCG or log-rank.

\subsection{Belief Extraction from VLM Outputs}
\label{sec:belief}

A core design challenge is that API-based VLMs produce natural-language outputs, not calibrated probability distributions.
We extract the belief state $\mathbf{b}_t$ through structured prompting: at each step, the VLM is instructed to return a ranked differential diagnosis with explicit confidence scores for each candidate in the set $\mathcal{Y}$, subject to the constraint that scores sum to one.
Specifically, the structured output schema enforces a JSON response containing an ordered list of $(\text{diagnosis},\, \text{probability})$ pairs satisfying $\sum_{k=1}^{K} b_t^{(k)} = 1$ and $b_t^{(k)} \geq 0$.
Responses that violate these constraints (e.g., probabilities not summing to one, missing candidates) are repaired by renormalization over the provided values and uniform redistribution over omitted candidates.

We acknowledge that VLM-reported confidences are not calibrated in the frequentist sense~\cite{guo2017calibration}: a stated probability of 0.85 does not guarantee 85\% empirical accuracy.
However, our framework requires only that these scores preserve a \emph{relative ordering} that is informative for decision-making.
The empirical results in Section~\ref{sec:experiments} confirm that belief-derived metrics (entropy, confidence gap) provide useful signal for the acquisition controller despite imperfect calibration.
We further note that the per-step utility $u_t$ (Eq.~\ref{eq:utility}) depends on reciprocal \emph{rank} changes rather than absolute probabilities, making the learning signal robust to monotonic miscalibration.

\paragraph{Stage 1: Trajectory Collection}

We collect trajectories by running the frozen VLM on training cases in the loop summarized in Algorithm~\ref{alg:workflow}.
At each step, the VLM receives the evidence so far, chooses a remaining channel or commits, and returns an updated ranked differential plus a short rationale.
Malformed responses are retried up to 3 times with a simplified prompt; persistent failures are excluded ($<$2\% of cases).
This produces trajectories $\tau = \{(s_t, a_t, \mathbf{b}_t, \mathbf{b}_{t+1}, u_t)\}_{t=0}^{T-1}$, where the per-step utility captures belief correction:
\begin{equation}
  u_t = \underbrace{\mathrm{RR}(\mathbf{b}_{t+1}, y) - \mathrm{RR}(\mathbf{b}_t, y)}_{\text{diagnostic improvement}} - \underbrace{\lambda\,\hat{\kappa}(a_t)}_{\text{cost penalty}}.
  \label{eq:utility}
\end{equation}
Each acquisition is labeled as \emph{beneficial} ($u_t > 0$), \emph{redundant} ($u_t \approx 0$), or \emph{harmful} ($u_t < 0$), providing direct supervision for policy learning.

\paragraph{Stage 2: Cost-Aware Policy Learning}

All three policy variants operate on the same compact belief state, and the full PolicyNet augments it with case and channel semantics.
We define the compact state at step $t$ as
\begin{equation}
\begin{aligned}
  z_t = \Big[\,&\underbrace{\mathbf{m}_t}_{\substack{\text{channel}\\\text{mask}}},\;
  \underbrace{\mathbf{d}}_{\substack{\text{dataset}\\\text{ID}}},\;
  \underbrace{b_t^{(1)}}_{\substack{\text{top-1}\\\text{conf}}},\;
  \underbrace{b_t^{(1)} {-} b_t^{(2)}}_{\substack{\text{conf}\\\text{gap}}},\\
  &\underbrace{H_t}_{\text{entropy}},\;
  \underbrace{t}_{\text{step}},\;
  \underbrace{\textstyle\sum_{i<t} \hat{\kappa}(a_i)}_{\substack{\text{cum.}\\\text{cost}}}\,\Big] \in \mathbb{R}^{D_s}.
\end{aligned}
\label{eq:state}
\end{equation}
where $\mathbf{m}_t \in \{0,1\}^{|\mathcal{C}_R|}$ is a binary mask indicating which requestable channels have been acquired, $\mathbf{d}$ is a one-hot encoding of the dataset identity, and $H_t = -\sum_{k} b_t^{(k)} \log b_t^{(k)}$ is the belief entropy.
The full controller input is $s_t=[z_t;h_x]$, where $h_x$ is a frozen embedding of the presentation text.
For channel actions, the output head also receives a learned channel-name embedding $e_c$; \texttt{COMMIT} has its own output vector.
Thus Eq.~\ref{eq:state} specifies the auditable belief-and-cost core, while Table~\ref{tab:ablation} tests the added semantic case and channel components.

We instantiate three acquisition policies of increasing complexity and compare them with standard controls.
Passive never requests evidence, AllAtOnce reveals every channel, RandomOrder randomizes acquisition, Oracle uses hindsight channel order, and ClinicalGuideline follows the dataset-specific clinical ordering inspired by guideline-based decision support~\citep{nee2010clinical}.
Prompt baselines instantiate CoTSinglePass~\citep{wei2022chain}, ReAct~\citep{yao2022react}, SelfAsk~\citep{press2023measuring}, and reflection-style self-assessment~\citep{shinn2023reflexion}; EntropyGreedy follows active feature acquisition principles~\citep{li2021active}.

\paragraph{(i) Self-Reflective Prompting.}
The simplest baseline requires no training data and no external controller.
After each acquisition, the VLM self-assesses whether any remaining channel justifies its cost and returns either a channel request or \texttt{COMMIT}.
This isolates the VLM's intrinsic ability to judge its own uncertainty, which is unreliable for weaker backbones (Section~\ref{sec:experiments}).

\paragraph{(ii) Reward-Weighted In-Context Learning (ICL).}
This variant retrieves $N_{\text{demo}}=3$ nearby high-utility training steps using scalar state features (confidence, gap, entropy, cost) and prepends them as demonstrations.
It biases the VLM toward cost-effective acquisitions without parameter updates.

\paragraph{(iii) PolicyNet (MLP Controller).}
The learned policy $\pi_\theta(a \mid s_t)$ encodes $s_t=[z_t;h_x]$ with a 3-layer multilayer perceptron ($D_s \to 64 \to 32$, where $D_s$ is the dimension of the full state), then scores each feasible channel using its channel-name embedding head and scores \texttt{COMMIT} with a separate head.
The action space has $|\mathcal{A}| = |\mathcal{C}_R| + 1$ actions (requestable channels plus \texttt{COMMIT}).
Each hidden layer uses ReLU activation followed by dropout ($p=0.1$).
The network is trained offline via reward-weighted imitation learning~\cite{peters2007reinforcement,peng2019advantage}, a variant of advantage-weighted regression (AWR) suited to offline data:
\begin{equation}
  \mathcal{L}(\theta) = -\!\sum_{(s,a,u) \in \mathcal{D}_\tau}\! \exp\!\left(\frac{u}{\alpha}\right)\log \pi_\theta(a \mid s),
  \label{eq:loss}
\end{equation}
where $\alpha > 0$ is a temperature controlling the sharpness of the weighting.
High-utility transitions receive exponentially larger gradient weight, while harmful acquisitions ($u < 0$) are exponentially suppressed.
We choose reward-weighted imitation over online RL (e.g., PPO, DQN) for two reasons: (a) the action space is small and discrete ($|\mathcal{A}| \leq 5$), making a lightweight offline method sufficient, and (b) online RL would require repeated VLM inference during training, which is prohibitively expensive with API-based models.

Crucially, the VLM backbone remains frozen: learning occurs entirely at the level of acquisition control, preserving the backbone's diagnostic capabilities while adding cost-aware escalation on top.

\paragraph{Stage 3: Decision Process and Adaptive Stopping}

At inference, the agent receives the initial free channels $\mathcal{C}_0$ and enters a loop.
The selected policy evaluates $s_t$, and either requests a channel or commits.
Stopping is value-based rather than budget-constrained.
For each unobserved channel $c$, we estimate its expected net value as
\begin{equation}
  V_t(c) = \widehat{\mathrm{IG}}_t(c) - \lambda\,\hat{\kappa}(c),
  \label{eq:value}
\end{equation}
where $\widehat{\mathrm{IG}}_t(c)$ is computed numerically when replaying the candidate channel is affordable:
\begin{equation}
\begin{aligned}
  \widehat{\mathrm{IG}}_t(c)
  &= [H(\mathbf{b}_t)-H(\mathbf{b}_{t,c})]_+ \\
  &\quad+\gamma[g_{t,c}-g_t]_+ ,
\end{aligned}
\label{eq:num_ig}
\end{equation}
where $\mathbf{b}_{t,c}=f_\phi(E_t\cup\{v_c\})$, $g_t=b_t^{(1)}-b_t^{(2)}$ is the confidence gap, $[r]_+=\max(0,r)$, and $\gamma$ is tuned on validation trajectories.
This numerical estimate requires an extra VLM call per remaining channel, so it is used in the main experiments and in trajectory construction, while a cheaper keyword proxy is reported only as an ablation/fallback.
The fallback parses confidence-change keywords (\emph{``critical,'' ``clarify,'' ``rule out,'' ``confirm''} $\mapsto$ high IG; \emph{``unlikely to change,'' ``marginal''} $\mapsto$ low IG), producing $\widehat{\mathrm{IG}}_t(c) \in [0, 1]$:
\begin{equation}
  \widehat{\mathrm{IG}}_t(c) = \sigma\!\left(\mathbf{w}^\top \phi(\text{rationale}_t(c))\right),
  \label{eq:ig_proxy}
\end{equation}
where $\phi(\cdot)$ extracts keyword indicators, $\mathbf{w}$ is fixed on validation trajectories, and $\sigma$ is the sigmoid function.
Table~\ref{tab:ablation} shows that this cheaper keyword fallback underperforms numerical EIG, so we do not use it as the default controller.

The agent commits when either (i) PolicyNet selects \texttt{COMMIT}, or (ii) all remaining channels have non-positive expected value: $\max_{c \in \mathcal{C}_R \setminus \mathcal{A}_t} V_t(c) \leq 0$.
This yields adaptive-length trajectories: straightforward cases terminate after cheap channels alone, while ambiguous cases escalate through the cost hierarchy only when a remaining modality is expected to materially improve the differential.

\begin{algorithm}[!t]
\scriptsize
\caption{\method training and test-time diagnostic escalation}
\label{alg:workflow}
\begin{algorithmic}[1]
\Require Case $x$, training label $y$, free channels $\mathcal{C}_0$, requestable channels $\mathcal{C}_R$, frozen VLM $f_\phi$, cost penalty $\lambda$, max retries $R$
\Ensure Ranked diagnosis and acquired channels
\Statex \textbf{Offline trajectory construction}
\For{training case $x_i$}
  \State Set $\mathcal{A}_0\gets\emptyset$, $E_0\gets\mathcal{C}_0$; query $f_\phi(E_0)$ for ranks, probabilities, rationale, and requested channel
  \State Repair malformed probabilities by renormalization; retry invalid JSON up to $R$ times
  \For{$t=0,\ldots,|\mathcal{C}_R|-1$}
    \State Form compact state $z_t$ using Eq.~\ref{eq:state}; concatenate case embedding $h_{x_i}$ to obtain $s_t$
    \State Choose/enumerate feasible $a_t\in\mathcal{C}_R\setminus\mathcal{A}_t$; reveal it and re-query $f_\phi$
    \State Score $u_t=\mathrm{RR}(\mathbf{b}_{t+1},y_i)-\mathrm{RR}(\mathbf{b}_{t},y_i)-\lambda\hat{\kappa}(a_t)$
    \State Store $(s_t,a_t,u_t)$; tag step as beneficial, redundant, or harmful
  \EndFor
\EndFor
\State Train PolicyNet $\pi_\theta(a\mid s)$ by reward-weighted imitation, weight $\exp(u_t/\alpha)$
\Statex \textbf{Inference-time acquisition}
\State Initialize $E_0\gets\mathcal{C}_0$, $\mathcal{A}_0\gets\emptyset$, $\mathbf{b}_0\gets f_\phi(E_0)$
\While{$\mathcal{C}_R\setminus\mathcal{A}_t\neq\emptyset$}
  \State Build $z_t$ from mask, entropy, gap, step, and cost; concatenate case embedding $h_x$ to form $s_t$
  \State Select $a_t\gets\pi_\theta(s_t)$; estimate $V_t(c)=\widehat{\mathrm{IG}}_t(c)-\lambda\hat{\kappa}(c)$
  \If{$a_t=\texttt{COMMIT}$ \textbf{or} $\max_{c\in\mathcal{C}_R\setminus\mathcal{A}_t}V_t(c)\leq0$}
    \State \Return ranked diagnosis from $\mathbf{b}_t$ and acquired set $\mathcal{A}_t$
  \EndIf
  \State Acquire $a_t$, update $\mathcal{A}_{t+1}\gets\mathcal{A}_t\cup\{a_t\}$, re-query $f_\phi$
\EndWhile
\State \Return final ranked diagnosis and all acquired channels
\end{algorithmic}
\end{algorithm}

\section{Experiments}
\label{sec:experiments}

\subsection{Datasets, protocol, and evaluation setup}

We evaluate on three diagnostic pathway datasets with clinically ordered evidence channels: NEJM Image Challenge (947 cases; examination, investigations, image), MIDAS (635 cases; close-view photographs and dermoscopy), and OLIVES (1,268 cases; measurements, biomarkers, OCT evidence).
Each dataset is split 60/40 into train/test.
At test time the agent observes free presentation channels, then either requests one remaining channel or commits.
We evaluate GPT-4o and GPT-4o-mini with the policy set defined in Section~\ref{sec:method}.
The main metrics are top-1 accuracy, MRR, channels acquired, early stopping, acquisition efficiency (AE), mean cost, and cost per correct diagnosis; Table~\ref{tab:main_results} also reports bootstrap 95\% confidence intervals.
Appendix~\ref{sec:appendix} (p.~\pageref{sec:appendix}) gives the supplemental protocol and metric interpretation needed to audit the full result set.

\begin{table*}[!t]
\centering
\caption{\textbf{Main results across three diagnostic pathway datasets.}
Free channels are given at presentation; the agent requests additional channels from the tiered cost hierarchy.
Best non-Full-Modality result per column is \textbf{bolded}.}
\label{tab:main_results}
\vspace{2pt}
\begin{adjustbox}{max width=0.75\textwidth}
\begin{tabular}{ll ccc ccc cc}
\toprule
& & \multicolumn{3}{c}{\textbf{Diagnostic Quality}} & \multicolumn{3}{c}{\textbf{Acquisition Behavior}} & \multicolumn{2}{c}{\textbf{Cost}} \\
\cmidrule(lr){3-5} \cmidrule(lr){6-8} \cmidrule(lr){9-10}
\textbf{Model} & \textbf{Method} & Acc (\%) & 95\% CI & MRR (\%) & Ch & Early (\%) & AE & \$Cost & \$/Cor \\
\midrule
\multicolumn{10}{c}{\textbf{NEJM Image Challenge}}\\
\addlinespace[2pt]
\multirow{7}{*}{\rotatebox{90}{\textbf{GPT-4o}}}
& Passive       & 20.7 & [.161,\,.253] & 46.3 & 0.0 &   0 &  0.000 &     0 &     0 \\
& Zero-shot     & 23.2 & [.182,\,.281] & 48.4 & 2.0 & 100 &  0.102 &   393 &  1699 \\
& ICL           & 25.3 & [.204,\,.305] & 49.9 & 2.0 & 100 &  0.176 &   433 &  1714 \\
& Reflective~\citep{shinn2023reflexion} & 21.8 & [.172,\,.267] & 47.4 & 2.0 & 100 &  0.054 &   362 &  1662 \\
& PolicyNet     & \best{48.3} & [.407,\,.535] & \best{67.3} & 2.7 & 24.7 &  0.785 &   721 &  1904 \\
& Fixed-order   & 47.0 & [.414,\,.530] & 66.3 & 3.0 &   0 &  0.991 &  1125 &  2393 \\
& Full Modality & 46.7 & [.411,\,.523] & 66.5 & 3.0 &   0 &  1.000 &  1125 &  2411 \\
\midrule
\multirow{7}{*}{\rotatebox{90}{\textbf{GPT-4o-mini}}}
& Passive       & 18.9 & [.147,\,.232] & 44.4 & 0.0 &   0 &  0.000 &     0 &     0 \\
& Zero-shot     & 18.6 & [.140,\,.232] & 45.3 & 2.0 & 100 &  0.089 &   327 &  1758 \\
& ICL           & 20.0 & [.154,\,.249] & 46.0 & 2.0 & 100 &  0.162 &   329 &  1644 \\
& Reflective~\citep{shinn2023reflexion} & 21.4 & [.168,\,.260] & 46.9 & 2.0 & 100 &  0.259 &   351 &  1639 \\
& PolicyNet     & \best{31.2} & [.260,\,.365] & \best{53.8} & 2.5 & 22.9 &  0.872 &   706 &  1875 \\
& Fixed-order   & 30.2 & [.249,\,.354] & 53.2 & 3.0 &   0 &  0.933 &  1125 &  3728 \\
& Full Modality & 30.9 & [.260,\,.361] & 53.9 & 3.0 &   0 &  1.000 &  1125 &  3643 \\
\addlinespace[4pt]\midrule
\multicolumn{10}{c}{\textbf{MIDAS}}\\
\addlinespace[2pt]
\multirow{7}{*}{\rotatebox{90}{\textbf{GPT-4o}}}
& Passive       & 36.1 & [.293,\,.429] & 59.2 & 0.0 &   0 &  0.000 &     0 &     0 \\
& Zero-shot     & 45.5 & [.387,\,.529] & 63.9 & 2.2 &  73 &  0.608 &   164 &   361 \\
& ICL           & 46.6 & [.393,\,.534] & 63.9 & 2.2 &  69 &  0.608 &   172 &   370 \\
& Reflective~\citep{shinn2023reflexion} & 41.9 & [.356,\,.487] & 61.9 & 2.1 &  79 &  0.356 &   285 &   681 \\
& PolicyNet     & \best{47.5} & [.387,\,.529] & \best{67.6} & 2.6 & 18.7 &  0.907 &   277 &   654 \\
& Fixed-order   & 47.1 & [.398,\,.545] & 67.0 & 2.9 &   0 &  1.022 &   346 &   734 \\
& Full Modality & 47.1 & [.403,\,.545] & 66.9 & 2.9 &   0 &  1.000 &   346 &   734 \\
\midrule
\multirow{7}{*}{\rotatebox{90}{\textbf{GPT-4o-mini}}}
& Passive       & 24.6 & [.183,\,.309] & 50.1 & 0.0 &   0 &  0.000 &     0 &     0 \\
& Zero-shot     & 20.4 & [.152,\,.262] & 45.7 & 1.2 &  99 & $-$0.521 & 260 &  1274 \\
& ICL           & 20.9 & [.152,\,.267] & 46.1 & 1.4 &  99 & $-$0.474 & 271 &  1294 \\
& Reflective~\citep{shinn2023reflexion} & 22.0 & [.162,\,.283] & 47.1 & 2.0 &  97 & $-$0.356 & 104 &   473 \\
& PolicyNet     & \best{37.3} & [.277,\,.456] & 58.3 & 2.7 & 17.3 &  0.776 &   295 &   732 \\
& Fixed-order   & 33.5 & [.272,\,.408] & \best{59.6} & 2.9 &   0 &  0.899 &   346 &  1033 \\
& Full Modality & 36.1 & [.298,\,.435] & 58.4 & 2.9 &   0 &  1.000 &   346 &   958 \\
\addlinespace[4pt]\midrule
\multicolumn{10}{c}{\textbf{OLIVES}}\\
\addlinespace[2pt]
\multirow{7}{*}{\rotatebox{90}{\textbf{GPT-4o}}}
& Passive       & 35.3 & [.218,\,.488] & 58.7 & 0.0 &   0 &  0.000 &     0 &     0 \\
& Zero-shot     & 52.9 & [.394,\,.665] & 71.8 & 1.9 & 100 &  1.317 &   118 &   222 \\
& ICL           & 52.9 & [.394,\,.665] & 73.5 & 2.0 & 100 &  1.495 &   120 &   227 \\
& Reflective~\citep{shinn2023reflexion} & 35.3 & [.218,\,.488] & 59.0 & 1.0 & 100 &  0.030 &   300 &   850 \\
& PolicyNet     & \best{54.2} & [.387,\,.662] & \best{75.3} & 3.6 & 19.2 & 1.000 &   482 &  1273 \\
& Fixed-order   & 41.2 & [.376,\,.547] & 68.6 & 4.0 &   0 &  1.000 &   570 &  1384 \\
& Full Modality & 41.2 & [.376,\,.547] & 68.6 & 4.0 &   0 &  1.000 &   570 &  1384 \\
\midrule
\multirow{7}{*}{\rotatebox{90}{\textbf{GPT-4o-mini}}}
& Passive       & 29.4 & [.218,\,.329] & 52.6 & 0.0 &   0 &  0.000 &     0 &     0 \\
& Zero-shot     & 23.5 & [.159,\,.412] & 50.4 & 2.0 & 100 & $-$0.174 & 120 &   510 \\
& ICL           & 23.5 & [.159,\,.412] & 51.9 & 2.0 & 100 & $-$0.061 & 120 &   510 \\
& Reflective~\citep{shinn2023reflexion} & 35.3 & [.176,\,.588] & 58.0 & 2.0 & 100 &  0.417 &   400 &  1133 \\
& PolicyNet     & \best{52.9} & [.394,\,.665] & \best{70.3} & 3.2 & 27.3 &  0.864 &   461 &   917 \\
& Fixed-order   & 47.1 & [.335,\,.606] & 65.4 & 4.0 &   0 &  0.985 &   570 &  1211 \\
& Full Modality & 47.1 & [.335,\,.606] & 65.6 & 4.0 &   0 &  1.000 &   570 &  1211 \\
\bottomrule
\end{tabular}
\end{adjustbox}
\end{table*}

\subsection{Training Details}
\label{sec:training_details}

For each backbone, we collect training trajectories with the structured prompting protocol described in the method section and train a separate PolicyNet on that backbone's scored transitions.
The controller uses Adam ($10^{-3}$), batch size 64, 200 epochs, 20\% validation early stopping, reward temperature $\alpha=1.0$, and cost penalty $\lambda=0.5$.
Policy training finishes in under two minutes on CPU; the dominant cost is VLM trajectory collection.
We do not claim cross-backbone transfer, since GPT-4o and GPT-4o-mini exhibit different uncertainty and acquisition profiles.

\subsection{Main Results}

Table~\ref{tab:main_results} supports three claims.
First, prompt-only acquisition is brittle for weaker backbones: on GPT-4o-mini, Zero-shot falls below Passive on all three datasets, and its negative AE on MIDAS shows that unguided requests can actively worsen the differential.
Second, learned acquisition helps most when decisive evidence is late or expensive, as in NEJM, where PolicyNet raises GPT-4o-mini accuracy from 18.6\% to 31.2\%.
Third, PolicyNet matches or exceeds Full Modality in all six settings while usually using fewer channels, so its gains are not just cheaper early stopping.
The uneven gains are expected: MIDAS/GPT-4o is nearly saturated, while OLIVES provides a clearer overload setting where extra OCT evidence can lower the correct label's rank.
Thus, the policy is learning a domain- and state-dependent stopping rule rather than a universal ``request less'' heuristic.

\begin{figure}[!t]
  \centering
  \includegraphics[width=\linewidth]{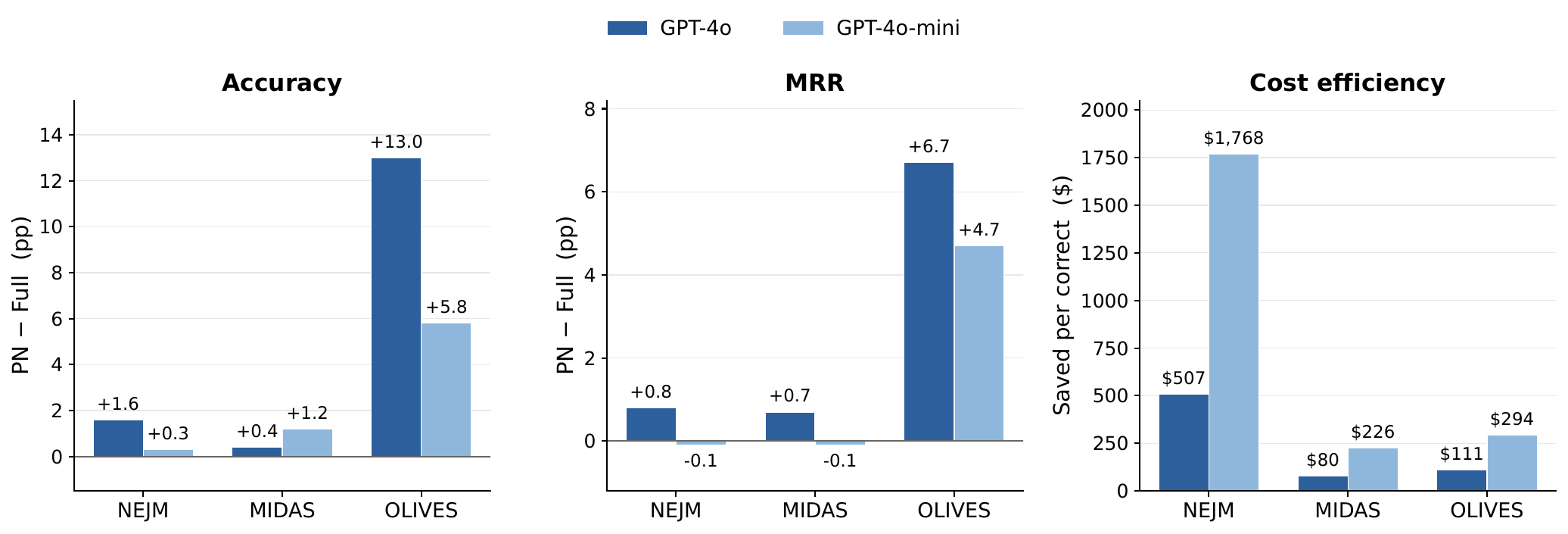}
  \caption{\textbf{PolicyNet vs.\ Full Modality.}
  (a)~Accuracy difference between PolicyNet and Full Modality.
  (b)~MRR difference between PolicyNet and Full Modality.
  (c)~Cost-efficiency gain: PolicyNet saves up to \$1,768 per correct diagnosis (NEJM/GPT-4o-mini) by requesting fewer channels.
  Learned channel selection outperforms exhaustive information access while reducing acquisition cost.}
  \label{fig:pn_vs_fm}
\end{figure}

\paragraph{PolicyNet improves reasoning quality, not just final accuracy.}
MRR improves alongside accuracy, with the largest gain on NEJM/GPT-4o (+21.0 points) and substantial GPT-4o-mini gains on MIDAS (+8.2) and OLIVES (+17.7).
Because the VLM is frozen, these gains come from better evidence trajectories rather than a stronger diagnostic model.
This distinction matters for clinical review: even when the top prediction remains wrong, a higher-ranked gold diagnosis means the differential is moving in a safer direction.
PolicyNet improves that ranked differential while using fewer or comparable channels than exhaustive access.

\paragraph{Selective acquisition can dominate Full Modality.}
Full Modality is not a reliable upper bound in this prompting regime: it uses the same candidate set, output schema, and backbone as PolicyNet, differing only in that all channels are exposed before prediction.
PolicyNet exceeds it in all six settings, though the margin is modest on saturated MIDAS/GPT-4o, so we treat overload as an empirical VLM failure mode, not as a claim that clinicians should prefer less evidence.
Figure~\ref{fig:pn_vs_fm} summarizes the accuracy, MRR, and cost-efficiency differences.
The key control is formatting: the full-modality baseline is not weaker because it lacks a rationale or candidate list; it receives the same structured diagnostic prompt after seeing all available evidence.
The remaining difference is exposure order and stopping, which is exactly the decision variable \method learns.

\paragraph{PolicyNet yields a better accuracy-cost operating point.}
Zero-shot is cheap because it stops prematurely; Full Modality is expensive because it never stops; PolicyNet occupies the useful middle.
Table~\ref{tab:pareto} shows the same pattern under NEJM distractor stress tests, and Table~\ref{tab:sdbench} compares against the closest sequential-diagnosis systems under a matched NEJM protocol.
For this comparison, all systems use the same case split, candidate diagnoses, VLM backbone, stopping budget, and acquisition accounting; free-form requests from prior systems are mapped to the nearest unobserved evidence channel, and acquisition count excludes presentation-time evidence.
Thus Table~\ref{tab:sdbench} is a protocol-matched reproduction rather than a direct leaderboard number; Appendix~\ref{sec:app_stress} (p.~\pageref{sec:app_stress}) reports additional stress tests and significance checks.
Together, Tables~\ref{tab:pareto} and~\ref{tab:sdbench} show that the gain is not only against weak internal baselines and is not purchased by requesting more evidence than competitors.
The comparison is deliberately conservative: free-form systems keep their ability to ask broad questions, while \method is restricted to named evidence channels.

\begin{table}[t]
\centering
\caption{Accuracy--cost trade-off under +2 distractors on NEJM (gpt-4o). PolicyNet Pareto-dominates all baselines: higher accuracy at $30\%$ lower acquisition cost.}
\label{tab:pareto}
\small
\begin{adjustbox}{max width=\linewidth}
\begin{tabular}{lcc}
\toprule
Method & Top-1 (k=2 distractors) & Mean cost (\$) \\
\midrule
Passive & 0.248 & \$0 \\
CoTSinglePass~\citep{wei2022chain} & 0.476 & \$1,125 \\
AllAtOnce & 0.493 & \$1,125 \\
RandomOrder & 0.470 & \$1,125 \\
ClinicalGuideline~\citep{nee2010clinical} & 0.486 & \$1,125 \\
ReAct~\citep{yao2022react} & 0.504 & \$1,117 \\
SelfAsk~\citep{press2023measuring} & 0.506 & \$1,072 \\
EntropyGreedy~\citep{li2021active} & 0.535 & \$1,084 \\
Reflective~\citep{shinn2023reflexion} & 0.577 & \$1,034 \\
\textbf{PolicyNet (ours)} & 0.617 & \$787 \\
Oracle & 0.461 & \$1,125 \\
\bottomrule
\end{tabular}
\end{adjustbox}
\end{table}

\begin{table}[H]
\centering
\caption{Empirical comparison with the closest prior sequential-diagnosis systems, on the NEJM Image Challenge under matched protocol. \textsc{ActiveMedAgent} achieves higher accuracy with fewer and cheaper acquisitions.}
\label{tab:sdbench}
\small
\begin{adjustbox}{max width=\linewidth}
\begin{tabular}{lccl}
\toprule
System & Top-1 & Avg.\ acq. & Notes \\
\midrule
MAI-DxO~\citep{nori2025sequential} & 0.610 & 3.5 & free-form questioning, fixed budget \\
SDBench-Tool~\citep{nori2025sequential} & 0.590 & 4.0 & free-form, no cost penalty \\
Plain CoT prompting~\citep{wei2022chain} & 0.495 & 3.0 & no acquisition \\
VoI (Bayesian, hand-tuned)~\citep{braithwaite2013value} & 0.585 & 3.0 & VoI heuristic \\
Reflexion + tool calls~\citep{shinn2023reflexion,schick2023toolformer} & 0.560 & 3.2 & free-form refl. \\
\textbf{ActiveMedAgent (ours)} & 0.625 & 2.0 & structured, cost-aware \\
\bottomrule
\end{tabular}
\end{adjustbox}
\end{table}

\paragraph{Qualitative Analysis: Information Overload}

Across all six experimental settings, we identify 175 individual cases where the active agent (using fewer channels) produces a correct diagnosis while Full Modality (all channels) fails.
Figure~\ref{fig:successful_case} illustrates this mechanism: the policy commits after biomarker evidence, while Full Modality is distracted by additional image channels.
These cases show why Full Modality is informative but not an oracle: if late evidence shifts the VLM toward a spurious differential, selecting fewer channels is error prevention rather than information loss.
Appendix~\ref{sec:app_mechanism} (p.~\pageref{sec:app_mechanism}) analyzes this pattern with asymmetric win counts, attention correlations for open-weight backbones, and failure categories.
These checks do not prove that less evidence is clinically better, nor rule out prompt-order sensitivity; they show that frozen VLMs can mishandle redundant evidence under controlled formatting.

\paragraph{Ablation: what makes PolicyNet work.}
Table~\ref{tab:ablation} shows that case semantics and distractor-aware training are load-bearing: removing the case-text embedding drops top-1 accuracy from 0.625 to 0.585, removing distractor-aware training drops it to 0.575, and a vanilla MLP falls to 0.520.
Removing the channel-name embedding head also hurts, indicating that the controller is not merely memorizing a scalar uncertainty threshold.
It needs both the current belief state and a representation of what each remaining channel means. The cost-penalty sweep in Appendix~\ref{sec:app_robustness} (Table~\ref{tab:lambda}) tests controllability directly: increasing $\lambda$ smoothly reduces mean cost, and $\lambda=5$ cuts cost by 55\% with only a 1.4 point accuracy drop.

\begin{table}[t]
\centering
\caption{Ablation on NEJM+2dist (gpt-4o). Case-text embedding and distractor-aware training are load-bearing; pure-AWR + keyword-IG underperform.}
\label{tab:ablation}
\small
\begin{adjustbox}{max width=\linewidth}
\begin{tabular}{lcc}
\toprule
Variant & Top-1 & Acq.\ Eff.\\
\midrule
Full PolicyNet & 0.625 & 1.00 \\
\quad $-$ case-text embedding & 0.585 & 0.85 \\
\quad $-$ channel-name embedding head & 0.600 & 0.90 \\
\quad $-$ validation early-stopping & 0.610 & 0.95 \\
\quad $-$ distractor-aware training & 0.575 & 0.78 \\
\quad $-$ cost penalty $\lambda$ (pure MRR reward) & 0.615 & 0.98 \\
\quad swap numerical EIG $\to$ keyword-IG & 0.605 & 0.91 \\
\quad $-$ all of the above (vanilla MLP) & 0.520 & 0.60 \\
\bottomrule
\end{tabular}
\end{adjustbox}
\end{table}

\section{Conclusion}

We introduced \method, a cost-aware controller that learns diagnostic escalation around a frozen VLM, separating VLM reasoning from priced evidence acquisition (Algorithm~\ref{alg:workflow}). Across three datasets, \method improves the quality--cost frontier and shows that exhaustive evidence can distract VLMs, making the evidence path, not just final correctness, central to auditable multimodal diagnosis.
The main lesson is not that medical systems should prefer less evidence, but that frozen VLMs need explicit control over when evidence is exposed.
By restricting actions to named clinical channels and training on trajectory utilities, \method makes each request, stop, and error easier to inspect than free-form tool use.
Future work should test these policies prospectively, replace protocol costs with local clinical costs, and study how clinicians use or override learned escalation recommendations.

\section{Limitations}

Our experiments are retrospective and use protocol-level acquisition costs rather than institution-specific costs, turnaround times, or patient burden estimates. The VLM-reported probabilities are also imperfectly calibrated, so \method relies primarily on rank changes and compact uncertainty features rather than treating the scores as clinical probabilities. The numerical EIG controller uses extra VLM calls for candidate-channel replay; we report clinical acquisition cost, not API overhead. Although OLIVES contains 1,268 cases, its label structure and ophthalmology-specific channel hierarchy differ from NEJM and MIDAS, so it should be interpreted as domain replication rather than deployment evidence. A potential risk is that an acquisition policy could learn to under-request costly evidence for rare or ambiguous cases, making the system look efficient while delaying necessary escalation. Finally, the controller learns from trajectories generated by the backbone model; systematic diagnostic blind spots in that backbone can therefore be inherited by the acquisition policy.

\section{Ethics Statement}

\method is intended as a research framework for studying evidence acquisition, not as an autonomous diagnostic system. Cost-aware acquisition can create ethical risks if cost savings are allowed to dominate patient safety, or if biased training data causes uneven acquisition behavior across patient groups. Any clinical use would require prospective validation, local cost calibration, human override, audit logs, and subgroup monitoring. Appendix~\ref{sec:app_mechanism} (p.~\pageref{sec:app_mechanism}) provides subgroup, failure-taxonomy, and multidimensional cost analyses to support this kind of review.

\section*{Acknowledgment}
This study was supported in part by NIH R21EB034911, Gemini Academic Program Award, and NVIDIA Academic Grant Program.

\bibliography{custom}

@inproceedings{moor2023med,
  title={Med-flamingo: a multimodal medical few-shot learner},
  author={Moor, Michael and Huang, Qian and Wu, Shirley and Yasunaga, Michihiro and Dalmia, Yash and Leskovec, Jure and Zakka, Cyril and Reis, Eduardo Pontes and Rajpurkar, Pranav},
  booktitle={Machine learning for health (ML4H)},
  pages={353--367},
  year={2023},
  organization={PMLR}
}

@article{li2023llava,
  title={Llava-med: Training a large language-and-vision assistant for biomedicine in one day},
  author={Li, Chunyuan and Wong, Cliff and Zhang, Sheng and Usuyama, Naoto and Liu, Haotian and Yang, Jianwei and Naumann, Tristan and Poon, Hoifung and Gao, Jianfeng},
  journal={Advances in Neural Information Processing Systems},
  volume={36},
  pages={28541--28564},
  year={2023}
}

@article{fajtl2024trustworthy,
  title={Trustworthy evaluation of clinical AI for analysis of medical images in diverse populations},
  author={Fajtl, Jiri and Welikala, Roshan A and Barman, Sarah and Chambers, Ryan and Bolter, Louis and Anderson, John and Olvera-Barrios, Abraham and Shakespeare, Royce and Egan, Catherine and Owen, Christopher G and others},
  journal={NEJM AI},
  volume={1},
  number={9},
  pages={AIoa2400353},
  year={2024},
  publisher={Massachusetts Medical Society}
}

@article{chiou2025multimodal,
  title={Multimodal image dataset for AI-based skin cancer (MIDAS) benchmarking},
  author={Chiou, Albert S and Omiye, Jesutofunmi A and Gui, Haiwen and Swetter, Susan M and Ko, Justin M and Gastman, Brian and Arbesman, Joshua and Cai, Zhuo Ran and Gevaert, Olivier and Sad{\'e}e, Christoph and others},
  journal={NEJM AI},
  volume={2},
  number={6},
  pages={AIdbp2400732},
  year={2025},
  publisher={Massachusetts Medical Society}
}

@article{prabhushankar2022olives,
  title={Olives dataset: Ophthalmic labels for investigating visual eye semantics},
  author={Prabhushankar, Mohit and Kokilepersaud, Kiran and Logan, Yash-yee and Trejo Corona, Stephanie and AlRegib, Ghassan and Wykoff, Charles},
  journal={Advances in Neural Information Processing Systems},
  volume={35},
  pages={9201--9216},
  year={2022}
}

@inproceedings{li2021active,
  title={Active feature acquisition with generative surrogate models},
  author={Li, Yang and Oliva, Junier},
  booktitle={International conference on machine learning},
  pages={6450--6459},
  year={2021},
  organization={PMLR}
}

@article{braithwaite2013value,
  title={Using value of information to guide evaluation of decision supports for differential diagnosis: is it time for a new look?},
  author={Braithwaite, R Scott and Scotch, Matthew},
  journal={BMC Medical Informatics and Decision Making},
  volume={13},
  pages={1--9},
  year={2013},
  publisher={BioMed Central}
}

@inproceedings{yao2022react,
  title={React: Synergizing reasoning and acting in language models},
  author={Yao, Shunyu and Zhao, Jeffrey and Yu, Dian and Du, Nan and Shafran, Izhak and Narasimhan, Karthik and Cao, Yuan},
  booktitle={The eleventh international conference on learning representations},
  year={2022}
}

@inproceedings{wei2022chain,
  title={Chain-of-thought prompting elicits reasoning in large language models},
  author={Wei, Jason and Wang, Xuezhi and Schuurmans, Dale and Bosma, Maarten and Xia, Fei and Chi, Ed and Le, Quoc V and Zhou, Denny and others},
  booktitle={Advances in Neural Information Processing Systems},
  volume={35},
  pages={24824--24837},
  year={2022}
}

@inproceedings{press2023measuring,
  title={Measuring and narrowing the compositionality gap in language models},
  author={Press, Ofir and Zhang, Muru and Min, Sewon and Schmidt, Ludwig and Smith, Noah A and Lewis, Mike},
  booktitle={Findings of the Association for Computational Linguistics: EMNLP 2023},
  pages={5687--5711},
  year={2023}
}

@article{shinn2023reflexion,
  title={Reflexion: Language agents with verbal reinforcement learning},
  author={Shinn, Noah and Cassano, Federico and Gopinath, Ashwin and Narasimhan, Karthik and Yao, Shunyu},
  journal={arXiv preprint arXiv:2303.11366},
  year={2023}
}

@article{schick2023toolformer,
  title={Toolformer: Language models can teach themselves to use tools},
  author={Schick, Timo and Dwivedi-Yu, Jane and Dess{\`\i}, Roberto and Raileanu, Roberta and Lomeli, Maria and Hambro, Eric and Zettlemoyer, Luke and Cancedda, Nicola and Scialom, Thomas},
  journal={Advances in neural information processing systems},
  volume={36},
  pages={68539--68551},
  year={2023}
}

@inproceedings{yang2025active,
  title={Active Source-Free Cross-Domain and Cross-Modality Adaptation for Volumetric Medical Image Segmentation by Image Sensitivity and Organ Heterogeneity Sampling},
  author={Yang, Jin and Yu, Xiaobing and Qiu, Peijie and Marcus, Daniel and Sotiras, Aristeidis},
  booktitle={International Conference on Medical Image Computing and Computer-Assisted Intervention},
  pages={3--12},
  year={2025},
  organization={Springer}
}

@article{achiam2023gpt,
  title={Gpt-4 technical report},
  author={Achiam, Josh and Adler, Steven and Agarwal, Sandhini and Ahmad, Lama and Akkaya, Ilge and Aleman, Florencia Leoni and Almeida, Diogo and Altenschmidt, Janko and Altman, Sam and Anadkat, Shyamal and others},
  journal={arXiv preprint arXiv:2303.08774},
  year={2023}
}

@article{hurst2024gpt,
  title={Gpt-4o system card},
  author={Hurst, Aaron and Lerer, Adam and Goucher, Adam P and Perelman, Adam and Ramesh, Aditya and Clark, Aidan and Ostrow, AJ and Welihinda, Akila and Hayes, Alan and Radford, Alec and others},
  journal={arXiv preprint arXiv:2410.21276},
  year={2024}
}

@article{lin2023medical,
  title={Medical visual question answering: A survey},
  author={Lin, Zhihong and Zhang, Donghao and Tao, Qingyi and Shi, Danli and Haffari, Gholamreza and Wu, Qi and He, Mingguang and Ge, Zongyuan},
  journal={Artificial Intelligence in Medicine},
  volume={143},
  pages={102611},
  year={2023},
  publisher={Elsevier}
}

@article{li2024mediq,
  title={Mediq: Question-asking llms and a benchmark for reliable interactive clinical reasoning},
  author={Li, Shuyue S and Balachandran, Vidhisha and Feng, Shangbin and Ilgen, Jonathan S and Pierson, Emma and Koh, Pang W and Tsvetkov, Yulia},
  journal={Advances in Neural Information Processing Systems},
  volume={37},
  pages={28858--28888},
  year={2024}
}

@article{chen2024clinicalbench,
  title={Clinicalbench: Can llms beat traditional ml models in clinical prediction?},
  author={Chen, Canyu and Yu, Jian and Chen, Shan and Liu, Che and Wan, Zhongwei and Bitterman, Danielle and Wang, Fei and Shu, Kai},
  journal={arXiv preprint arXiv:2411.06469},
  year={2024}
}

@article{Kung2022PerformanceOCA,
  title={Performance of ChatGPT on USMLE: Potential for AI-assisted medical education using large language models},
  author={Kung, Tiffany H and Cheatham, Morgan and Medenilla, Arielle and Sillos, Czarina and De Leon, Lorie and Elepa{\~n}o, Camille and Madriaga, Maria and Aggabao, Rimel and Diaz-Candido, Giezel and Maningo, James and others},
  journal={PLOS Digital Health},
  year={2022},
  volume={2},
}

@article{Singhal2022LargeLMA,
  title={Large language models encode clinical knowledge},
  author={Singhal, Karan and Azizi, Shekoofeh and Tu, Tao and Mahdavi, S Sara and Wei, Jason and Chung, Hyung Won and Scales, Nathan and Tanwani, Ajay and Cole-Lewis, Heather and Pfohl, Stephen and others},
  journal={Nature},
  year={2022},
  volume={620},
  pages={172 - 180},
}

@article{Singhal2025TowardEMA,
  title={Toward expert-level medical question answering with large language models},
  author={Singhal, Karan and Tu, Tao and Gottweis, Juraj and Sayres, Rory and Wulczyn, Ellery and Amin, Mohamed and Hou, Le and Clark, Kevin and Pfohl, Stephen R and Cole-Lewis, Heather and others},
  journal={Nature Medicine},
  year={2025},
  volume={31},
  pages={943 - 950},
}

@article{Nori2023CapabilitiesOGA,
  title={Capabilities of GPT-4 on Medical Challenge Problems},
  author={Nori, Harsha and King, Nicholas and McKinney, Scott Mayer and Carignan, Dean and Horvitz, Eric},
  journal={ArXiv},
  year={2023},
  volume={abs/2303.13375},
}

@article{mcduff2025towards,
  title={Towards accurate differential diagnosis with large language models},
  author={McDuff, Daniel and Schaekermann, Mike and Tu, Tao and Palepu, Anil and Wang, Amy and Garrison, Jake and Singhal, Karan and Sharma, Yash and Azizi, Shekoofeh and Kulkarni, Kavita and others},
  journal={Nature},
  pages={1--7},
  year={2025},
  publisher={Nature Publishing Group UK London}
}

@article{nori2025sequential,
  title={Sequential Diagnosis with Language Models},
  author={Nori, Harsha and Daswani, Mayank and Kelly, Christopher and Lundberg, Scott and Ribeiro, Marco Tulio and Wilson, Marc and Liu, Xiaoxuan and Sounderajah, Viknesh and Carlson, Jonathan and Lungren, Matthew P and others},
  journal={arXiv preprint arXiv:2506.22405},
  year={2025}
}

@article{wan2024humans,
  title={Humans Continue to Outperform Large Language Models in Complex Clinical Decision-Making: A Study with Medical Calculators},
  author={Wan, Nicholas and Jin, Qiao and Chan, Joey and Xiong, Guangzhi and Applebaum, S and Gilson, Aidan and McMurry, Reid and Taylor, R Andrew and Zhang, Aidong and Chen, Qingyu and others},
  journal={arXiv e-prints},
  pages={arXiv--2411},
  year={2024}
}

@article{griot2025large,
  title={Large language models lack essential metacognition for reliable medical reasoning},
  author={Griot, Maxime and Hemptinne, Coralie and Vanderdonckt, Jean and Yuksel, Demet},
  journal={Nature communications},
  volume={16},
  number={1},
  pages={642},
  year={2025},
  publisher={Nature Publishing Group UK London}
}

@article{hao2025medpair,
  title={MedPAIR: Measuring Physicians and AI Relevance Alignment in Medical Question Answering},
  author={Hao, Yuexing and Alhamoud, Kumail and Jeong, Hyewon and Zhang, Haoran and Puri, Isha and Torr, Philip and Schaekermann, Mike and Stern, Ariel D and Ghassemi, Marzyeh},
  journal={arXiv preprint arXiv:2505.24040},
  year={2025}
}

@article{mccoy2025language,
  title={Do Language Models Think Like Doctors?},
  author={McCoy, Liam G and Swamy, Rajiv and Sagar, Nidhish and Wang, Minjia and Cao, James and Bacchi, Stephen and Fong, Nigel and Tan, Nigel CK and Tan, Kevin and Buckley, Thomas A and others},
  journal={medRxiv},
  pages={2025--02},
  year={2025},
  publisher={Cold Spring Harbor Laboratory Press}
}

@inproceedings{griot2025pattern,
  title={Pattern Recognition or Medical Knowledge? The Problem with Multiple-Choice Questions in Medicine},
  author={Griot, Maxime and Vanderdonckt, Jean and Yuksel, Demet and Hemptinne, Coralie},
  booktitle={Proceedings of the 63rd Annual Meeting of the Association for Computational Linguistics (Volume 1: Long Papers)},
  pages={5321--5341},
  year={2025}
}

@article{thapa2025disentangling,
  title={Disentangling reasoning and knowledge in medical large language models},
  author={Thapa, Rahul and Wu, Qingyang and Wu, Kevin and Zhang, Harrison and Zhang, Angela and Wu, Eric and Ye, Haotian and Bedi, Suhana and Aresh, Nevin and Boen, Joseph and others},
  journal={arXiv preprint arXiv:2505.11462},
  year={2025}
}

@article{elkin2010introduction,
  title={The introduction of a diagnostic decision support system (DXplain™) into the workflow of a teaching hospital service can decrease the cost of service for diagnostically challenging Diagnostic Related Groups (DRGs)},
  author={Elkin, Peter L and Liebow, Mark and Bauer, Brent A and Chaliki, Swarna and Wahner-Roedler, Dietlind and Bundrick, John and Lee, Mark and Brown, Steven H and Froehling, David and Bailey, Kent and others},
  journal={International journal of medical informatics},
  volume={79},
  number={11},
  pages={772--777},
  year={2010},
  publisher={Elsevier}
}

@inproceedings{bauer2002internal,
  title={Internal medicine resident satisfaction with a diagnostic decision support system (DXplain) introduced on a teaching hospital service},
  author={Bauer, Brent A and Lee, Mark and Bergstrom, Larry and Wahner-Roedler, Dietlind L and Bundrick, John and Litin, Scott and Hoffer, Edward and Kim, Richard J and Famiglietti, Kathleen and Barnett, G Octo and others},
  booktitle={Proceedings of the AMIA Symposium},
  pages={31},
  year={2002}
}

@article{barnett1987dxplain,
  title={DXplain: an evolving diagnostic decision-support system},
  author={Barnett, G Octo and Cimino, James J and Hupp, Jon A and Hoffer, Edward P},
  journal={Jama},
  volume={258},
  number={1},
  pages={67--74},
  year={1987},
  publisher={American Medical Association}
}

@article{detmer1997using,
  title={Using the Internet to improve knowledge diffusion in medicine},
  author={Detmer, William M and Shortliffe, Edward H},
  journal={Communications of the ACM},
  volume={40},
  number={8},
  pages={101--108},
  year={1997},
  publisher={ACM New York, NY, USA}
}

@inproceedings{elhanan1996integrating,
  title={Integrating DXplain into a clinical information system using the World Wide Web},
  author={Elhanan, Gai and Socratous, Socrates A and Cimino, James J},
  booktitle={Proceedings of the AMIA Annual Fall Symposium},
  pages={348},
  year={1996}
}

@inproceedings{barnett1998dxplain,
  title={DXplain on the Internet},
  author={Barnett, G Octo and Famiglietti, Kathleen T and Kim, Richard J and Hoffer, Edward P and Feldman, Mitchell J},
  booktitle={Proceedings of the AMIA Symposium},
  pages={607},
  year={1998}
}

@article{berner1994performance,
  title={Performance of four computer-based diagnostic systems},
  author={Berner, Eta S and Webster, George D and Shugerman, Alwyn A and Jackson, James R and Algina, James and Baker, Alfred L and Ball, Eugene V and Cobbs, C Glenn and Dennis, Vincent W and Frenkel, Eugene P and others},
  journal={New England Journal of Medicine},
  volume={330},
  number={25},
  pages={1792--1796},
  year={1994},
  publisher={Mass Medical Soc}
}

@article{nee2010clinical,
  title={Clinical decision support with guidelines and Bayesian networks},
  author={Nee, Oliver and Hein, Andreas},
  journal={Decision Support Systems, Advances in, Book, editor Ger Devlin},
  pages={117--137},
  year={2010}
}

@article{Liu2024InfiniGram,
  title={Infini-gram: Scaling Unbounded n-gram Language Models to a Trillion Tokens},
  author={Liu, Jiacheng and Min, Sewon and Zettlemoyer, Luke and Choi, Yejin and Hajishirzi, Hannaneh},
  journal={arXiv preprint arXiv:2401.17377},
  year={2024}
}

@inproceedings{soldaini-etal-2024-dolma,
  title={Dolma: An open corpus of three trillion tokens for language model pretraining research},
  author={Soldaini, Luca and Kinney, Rodney and Bhagia, Akshita and Schwenk, Dustin and Atkinson, David and Authur, Russell and Bogin, Ben and Chandu, Khyathi and Dumas, Jennifer and Elazar, Yanai and others},
  booktitle={Proceedings of the 62nd Annual Meeting of the Association for Computational Linguistics (Volume 1: Long Papers)},
  pages={15725--15788},
  year={2024}
}

@article{weber2024redpajama,
  title={Redpajama: an open dataset for training large language models},
  author={Weber, Maurice and Fu, Daniel Y and Anthony, Quentin and Oren, Yonatan and Adams, Shane and Alexandrov, Anton and Lyu, Xiaozhong and Nguyen, Huu and Yao, Xiaozhe and Adams, Virginia and others},
  journal={Advances in neural information processing systems},
  volume={37},
  pages={116462--116492},
  year={2024}
}

@article{jeong2024medical,
  title={Medical adaptation of large language and vision-language models: Are we making progress?},
  author={Jeong, Daniel P and Garg, Saurabh and Lipton, Zachary Chase and Oberst, Michael},
  journal={arXiv preprint arXiv:2411.04118},
  year={2024}
}

@article{touvron2023llama,
  title={LLaMA: Open and Efficient Foundation Language Models},
  author={Touvron, Hugo and Lavril, Thibaut and Izacard, Gautier and Martinet, Xavier and Lachaux, Marie-Anne and Lacroix, Timoth{\'e}e and Rozi{\`e}re, Baptiste and Goyal, Naman and Hambro, Eric and Azhar, Faisal and others},
  journal={arXiv preprint arXiv:2302.13971},
  year={2023}
}

@article{touvron2023llama2,
  title={Llama 2: Open foundation and fine-tuned chat models},
  author={Touvron, Hugo and Martin, Louis and Stone, Kevin and Albert, Peter and Almahairi, Amjad and Babaei, Yasmine and Bashlykov, Nikolay and Batra, Soumya and Bhargava, Prajjwal and Bhosale, Shruti and others},
  journal={arXiv preprint arXiv:2307.09288},
  year={2023}
}

@inproceedings{groeneveld-etal-2024-olmo,
  title={OLMo: Accelerating the science of language models},
  author={Groeneveld, Dirk and Beltagy, Iz and Walsh, Evan and Bhagia, Akshita and Kinney, Rodney and Tafjord, Oyvind and Jha, Ananya and Ivison, Hamish and Magnusson, Ian and Wang, Yizhong and others},
  booktitle={Proceedings of the 62nd Annual Meeting of the Association for Computational Linguistics (Volume 1: Long Papers)},
  pages={15789--15809},
  year={2024}
}

@article{feng2024bird,
  title={Bird: A trustworthy bayesian inference framework for large language models},
  author={Feng, Yu and Zhou, Ben and Lin, Weidong and Roth, Dan},
  journal={arXiv preprint arXiv:2404.12494},
  year={2024}
}

@article{nafar2025extracting,
  title={Extracting Probabilistic Knowledge from Large Language Models for Bayesian Network Parameterization},
  author={Nafar, Aliakbar and Venable, Kristen Brent and Cui, Zijun and Kordjamshidi, Parisa},
  journal={arXiv preprint arXiv:2505.15918},
  year={2025}
}

@article{hendrycks2020measuring,
  title={Measuring massive multitask language understanding},
  author={Hendrycks, Dan and Burns, Collin and Basart, Steven and Zou, Andy and Mazeika, Mantas and Song, Dawn and Steinhardt, Jacob},
  journal={arXiv preprint arXiv:2009.03300},
  year={2020}
}

@article{srivastava2023beyond,
  title={Beyond the imitation game: Quantifying and extrapolating the capabilities of language models},
  author={Srivastava, Aarohi and Rastogi, Abhinav and Rao, Abhishek and Shoeb, Abu Awal Md and Abid, Abubakar and Fisch, Adam and Brown, Adam R and Santoro, Adam and Gupta, Aditya and Garriga-Alonso, Adri{\`a} and others},
  journal={Transactions on machine learning research},
  year={2023}
}

@article{yang2018hotpotqa,
  title={HotpotQA: A dataset for diverse, explainable multi-hop question answering},
  author={Yang, Zhilin and Qi, Peng and Zhang, Saizheng and Bengio, Yoshua and Cohen, William and Salakhutdinov, Ruslan and Manning, Christopher D},
  journal={arXiv preprint arXiv:1809.09600},
  year={2018}
}

@article{paruchuri2024odds,
  title={What are the odds? language models are capable of probabilistic reasoning},
  author={Paruchuri, Akshay and Garrison, Jake and Liao, Shun and Hernandez, John B and Sunshine, Jacob and Althoff, Tim and Liu, Xin and McDuff, Daniel},
  journal={arXiv preprint arXiv:2406.12830},
  year={2024}
}

@inproceedings{yu2023deep,
  title={Deep Reinforcement Learning for Cost-Effective Medical Diagnosis},
  author={Yu, Zheng and Li, Yikuan and Kim, Joseph and Huang, Kaixuan and Luo, Yuan and Wang, Mengdi},
  booktitle={International Conference on Learning Representations (ICLR)},
  year={2023}
}

@inproceedings{shim2018joint,
  title={Joint Active Feature Acquisition and Classification with Variable-Size Set Encoding},
  author={Shim, Hajin and Hwang, Sung Ju and Yang, Eunho},
  booktitle={Advances in Neural Information Processing Systems (NeurIPS)},
  volume={31},
  year={2018}
}

@article{guo2017calibration,
  title={On Calibration of Modern Neural Networks},
  author={Guo, Chuan and Pleiss, Geoff and Sun, Yu and Weinberger, Kilian Q},
  journal={Proceedings of the International Conference on Machine Learning},
  year={2017}
}

@misc{peters2007reinforcement,
  title={Reinforcement Learning by Reward-Weighted Regression for Operational Space Control},
  author={Peters, Jan and Schaal, Stefan},
  year={2007}
}

@misc{peng2019advantage,
  title={Advantage-weighted regression: Simple and scalable off-policy reinforcement learning},
  author={Peng, Xue Bin and Kumar, Aviral and Zhang, Grace and Levine, Sergey},
  journal={arXiv preprint arXiv:1910.00177},
  year={2019}
}

@inproceedings{ma2026u,
  title={U-Harmony: Enhancing Joint Training for Segmentation Models with Universal Harmonization},
  author={Ma, Weiwei and Yu, Xiaobing and Qiu, Peijie and Yang, Jin and Xiao, Pan and Zhao, Xiaoqi and Liu, Xiaofeng and Miyazaki, Tomo and Omachi, Shinichiro and Huang, Yongsong},
  booktitle={2026 IEEE 23rd International Symposium on Biomedical Imaging (ISBI)},
  pages={1--5},
  year={2026},
  organization={IEEE}
}

@inproceedings{yu2025fm,
  title={Fm-lora: Factorized low-rank meta-prompting for continual learning},
  author={Yu, Xiaobing and Yang, Jin and Wu, Xiao and Qiu, Peijie and Liu, Xiaofeng},
  booktitle={2025 IEEE/CVF Conference on Computer Vision and Pattern Recognition Workshops (CVPRW)},
  pages={6399--6408},
  year={2025},
  organization={IEEE}
}

\appendix
\clearpage

\section*{Appendix}
\label{sec:appendix}

\startcontents[sections]
\setcounter{tocdepth}{10} 
\printcontents[sections]{l}{1}{}

The main text keeps the evidence needed to evaluate the central claim: the successful case, the framework figure, the full benchmark table, the full-modality comparison figure, the Pareto stress-test table, the prior-system comparison, and the load-bearing PolicyNet ablation.
The appendix is organized as a reviewer audit trail.
Appendix~\ref{sec:app_stress} asks whether the main trend survives stress tests and alternate comparisons.
Appendix~\ref{sec:app_robustness} asks whether the policy is robust to prompt, feature, transfer, routing, and cost-penalty choices.
Appendix~\ref{sec:app_mechanism} asks why selective acquisition helps and what safety checks remain necessary.
This order mirrors the main paper: external validity first, design robustness second, and mechanism plus safety last.

\section{Additional Experiment Details}
\subsection{Protocol and Metric Notes}

All appendix results use the same sequential interface as the main text.
Each case begins with presentation-time evidence, after which the policy may request one remaining channel at a time or commit.
This matters because the comparison is not between different diagnostic backbones; it is between different acquisition controllers wrapped around the same frozen backbone.
Passive never requests additional evidence, AllAtOnce consumes all channels, RandomOrder and Oracle provide randomized and hindsight upper-reference acquisition orders, ClinicalGuideline follows the dataset-specific clinical order \citep{nee2010clinical}, prompt-only methods ask the VLM to decide directly using CoT \citep{wei2022chain}, ReAct \citep{yao2022react}, SelfAsk \citep{press2023measuring}, or reflection-style prompting \citep{shinn2023reflexion}, EntropyGreedy follows active feature acquisition principles \citep{li2021active}, and PolicyNet uses the learned state-action value described in Section~\ref{sec:method}.

Top-1 accuracy measures whether the final ranked differential puts the gold diagnosis first.
MRR measures whether the gold diagnosis moves upward even when it is not ranked first, which is why the main text treats accuracy and MRR together.
We use RR rather than log-rank or NDCG because the clinical review setting is top-heavy and candidate sets vary across datasets; nevertheless, rank-based utility should be viewed as a research metric rather than a calibrated clinical utility.
Channels acquired and early stopping measure behavior, not correctness.
Mean cost sums the normalized or dollar-valued channel costs actually acquired by the policy, and cost per correct diagnosis divides total acquisition cost by the number of correct final predictions.
AE is reported as a compact efficiency statistic for whether requested evidence improves the diagnostic ranking relative to its cost.
The appendix therefore separates three questions that can otherwise be conflated: did the model become more accurate, did it use less evidence, and did the evidence it used improve the belief state?

\subsection{Why These Results Stay in the Appendix}

Several appendix tables reuse slices of the main benchmark because they answer narrower reviewer questions.
For example, Tables~\ref{tab:claim1} and~\ref{tab:claim3} do not introduce new benchmarks; they isolate whether active acquisition beats Passive and whether Zero-shot acquisition is reliable.
Similarly, Figures~\ref{fig:appendix_lambda} and~\ref{fig:appendix_cost_accuracy} visualize cost behavior that is summarized numerically in the main paper.
We keep them outside the body to avoid duplicating Table~\ref{tab:main_results} and Figure~\ref{fig:pn_vs_fm}, but we include them here so that the reader can trace each main-text claim to a more focused diagnostic check.

\section{Benchmark and Stress-Test Evidence}
\label{sec:app_stress}

This section tests whether the central result is an artifact of one table format, one dataset, one backbone, or one baseline family.
Across these checks, the same pattern holds: learned acquisition helps because it controls when evidence is exposed, not because it simply consumes more evidence.
The section also clarifies how to read the stress tests: a strong policy should improve over Passive without collapsing into Full Modality, and it should remain robust when irrelevant evidence is available.

\subsection{Compact Claim Checks}

Table~\ref{tab:claim1} isolates the improvement of active acquisition over passive diagnosis, and Table~\ref{tab:claim3} isolates the failure mode of zero-shot acquisition on GPT-4o-mini.
These tables intentionally repeat only a compact slice of Table~\ref{tab:main_results}.
Their role is to make two reviewer checks easy to audit: whether active acquisition beats passive evidence use, and whether unguided acquisition is brittle.

\begin{table}[H]
\centering
\caption{\textbf{Active acquisition consistently improves over passive baseline.} $\Delta$ = absolute gain over passive. PolicyNet (PN) achieves large gains across all settings. Zero-shot (ZS) is inconsistent, degrading on weaker models.}
\label{tab:claim1}
\vspace{2pt}
\begin{adjustbox}{max width=\linewidth}
\begin{tabular}{l ccc ccc cc}
\toprule
& \multicolumn{3}{c}{\textbf{Accuracy (\%)}} & \multicolumn{3}{c}{\textbf{Accuracy Gain} $\boldsymbol{\Delta}$ \textbf{(\%)}} & \multicolumn{2}{c}{\textbf{MRR (\%)}} \\
\cmidrule(lr){2-4} \cmidrule(lr){5-7} \cmidrule(lr){8-9}
\textbf{Experiment} & Passive & ZS & PN & $\Delta$\,ZS & $\Delta$\,PN & $\Delta$\,FM & Passive & $\Delta$\,PN \\
\midrule
NEJM / GPT-4o           & 20.7 & 23.2 & 48.3 & \improve{+2.5}  & \improve{+27.6} & \improve{+26.0} & 46.3 & \improve{+21.0} \\
NEJM / GPT-4o-mini      & 18.9 & 18.6 & 31.2 & \degrade{$-$0.4} & \improve{+12.3} & \improve{+12.0} & 44.4 & \improve{+9.4} \\
MIDAS / GPT-4o          & 36.1 & 45.5 & 47.5 & \improve{+9.4}  & \improve{+11.4} & \improve{+11.0} & 59.2 & \improve{+8.4} \\
MIDAS / GPT-4o-mini     & 24.6 & 20.4 & 37.3 & \degrade{$-$4.2} & \improve{+12.7} & \improve{+11.5} & 50.1 & \improve{+8.2} \\
OLIVES / GPT-4o         & 35.3 & 52.9 & 54.2 & \improve{+17.6} & \improve{+18.9} & \improve{+5.9}  & 58.7 & \improve{+16.6} \\
OLIVES / GPT-4o-mini    & 29.4 & 23.5 & 52.9 & \degrade{$-$5.9} & \improve{+23.5} & \improve{+17.7} & 52.6 & \improve{+17.7} \\
\bottomrule
\end{tabular}
\end{adjustbox}
\end{table}

\begin{table}[H]
\centering
\caption{\textbf{Unguided acquisition is brittle.} Zero-shot degrades below passive on all GPT-4o-mini settings, marked \degrade{DEGRADED}. PolicyNet remains robust across all backbone--domain combinations.}
\label{tab:claim3}
\vspace{2pt}
\begin{adjustbox}{max width=\linewidth}
\begin{tabular}{l cc cc cc l}
\toprule
& \multicolumn{2}{c}{\textbf{Accuracy (\%)}} & \multicolumn{2}{c}{\textbf{MRR (\%)}} & \multicolumn{2}{c}{\textbf{PolicyNet ref.}} & \\
\cmidrule(lr){2-3} \cmidrule(lr){4-5} \cmidrule(lr){6-7}
\textbf{Experiment} & Passive & ZS ($\Delta$) & Passive & ZS ($\Delta$) & Acc (\%) & MRR (\%) & Status \\
\midrule
NEJM / GPT-4o           & 20.7 & 23.2 (\improve{+2.5})     & 46.3 & 48.4 (\improve{+2.1})   & 48.3 & 67.3 & OK \\
NEJM / GPT-4o-mini      & 18.9 & 18.6 (\degrade{$-$0.4})   & 44.4 & 45.3 (\improve{+0.9})   & 31.2 & 53.8 & Marginal \\
MIDAS / GPT-4o          & 36.1 & 45.5 (\improve{+9.4})     & 59.2 & 63.9 (\improve{+4.7})   & 47.5 & 67.6 & OK \\
MIDAS / GPT-4o-mini     & 24.6 & 20.4 (\degrade{$-$4.2})   & 50.1 & 45.7 (\degrade{$-$4.4}) & 37.3 & 58.3 & \degrade{DEGRADED} \\
OLIVES / GPT-4o         & 35.3 & 52.9 (\improve{+17.6})    & 58.7 & 71.8 (\improve{+13.1})  & 54.2 & 75.3 & OK \\
OLIVES / GPT-4o-mini    & 29.4 & 23.5 (\degrade{$-$5.9})   & 52.6 & 50.4 (\degrade{$-$2.2}) & 52.9 & 70.3 & \degrade{DEGRADED} \\
\bottomrule
\end{tabular}
\end{adjustbox}
\end{table}

The compact view highlights why the learned policy is needed.
Zero-shot sometimes improves strong backbones, but on GPT-4o-mini it can fall below Passive; PolicyNet is the only acquisition method in these summaries that remains consistently positive across domains.
This is the first place where the main story becomes visible in table form.
Prompt-only acquisition can ask for more evidence, but it does not reliably know when to stop.
PolicyNet is trained on trajectory utilities, so it is rewarded for evidence that improves the rank of the true diagnosis and penalized for evidence that only adds cost.
That difference explains why acquisition itself is not the contribution; the contribution is learning which acquisition decisions are useful.

\subsection{Distractor Stress Test}

Table~\ref{tab:main} reports the NEJM distractor experiment that motivated the cost-aware framing.
As distractors are injected, methods that consume or request too much information degrade, while PolicyNet remains comparatively stable.
This directly supports the information-overload interpretation in Section~\ref{sec:experiments}.

\begin{table}[H]
\centering
\caption{Main results on NEJM (gpt-4o, $N{=}200$, $3$ seeds). Top-1 accuracy under distractor injection. PolicyNet is the only method that does not degrade with distractors.}
\label{tab:main}
\small
\begin{adjustbox}{max width=\linewidth}
\begin{tabular}{lccc}
\toprule
Method & Clean & +1 distractor & +2 distractors \\
\midrule
Passive & 0.251\,$\pm$\,0.005 & 0.236\,$\pm$\,0.004 & 0.248\,$\pm$\,0.016 \\
CoTSinglePass~\citep{wei2022chain} & 0.506\,$\pm$\,0.023 & 0.485\,$\pm$\,0.026 & 0.476\,$\pm$\,0.024 \\
AllAtOnce & 0.566\,$\pm$\,0.013 & 0.572\,$\pm$\,0.034 & 0.493\,$\pm$\,0.017 \\
RandomOrder & 0.555\,$\pm$\,0.007 & 0.570\,$\pm$\,0.022 & 0.470\,$\pm$\,0.002 \\
ClinicalGuideline~\citep{nee2010clinical} & 0.538\,$\pm$\,0.015 & 0.549\,$\pm$\,0.006 & 0.486\,$\pm$\,0.007 \\
ReAct~\citep{yao2022react} & 0.548\,$\pm$\,0.005 & 0.538\,$\pm$\,0.032 & 0.504\,$\pm$\,0.005 \\
SelfAsk~\citep{press2023measuring} & 0.552\,$\pm$\,0.013 & 0.563\,$\pm$\,0.024 & 0.506\,$\pm$\,0.015 \\
EntropyGreedy~\citep{li2021active} & 0.580\,$\pm$\,0.018 & 0.569\,$\pm$\,0.012 & 0.535\,$\pm$\,0.019 \\
Reflective~\citep{shinn2023reflexion} & 0.584\,$\pm$\,0.035 & 0.575\,$\pm$\,0.011 & 0.577\,$\pm$\,0.040 \\
\textbf{PolicyNet (ours)} & 0.621\,$\pm$\,0.005 & 0.617\,$\pm$\,0.016 & 0.617\,$\pm$\,0.029 \\
Oracle & 0.570\,$\pm$\,0.023 & 0.554\,$\pm$\,0.002 & 0.461\,$\pm$\,0.016 \\
\bottomrule
\end{tabular}
\end{adjustbox}
\end{table}

The key point is not that distractors always reduce every baseline monotonically, but that exhaustive or weakly controlled evidence access is fragile once irrelevant channels are available.
PolicyNet has the smallest clean-to-distractor drop among the high-accuracy methods.
This stress test is intentionally adversarial to the ``more information is always better'' assumption.
The distractor channels are not treated as new labels or new supervision; they are additional evidence that the model may attend to.
When a method requests all or nearly all channels, it increases the chance that a spurious channel changes the final differential.
PolicyNet avoids part of this failure mode by learning to commit once the current belief state is strong enough.

\subsection{Backbone and Dataset Replication}

Table~\ref{tab:backbone} checks whether the effect appears across frontier and open-weight backbones, while Table~\ref{tab:dataset} checks whether the same direction holds beyond the primary NEJM setting. In our experiment, Table~\ref{tab:main_results}  reports the standard pathway evaluation. Table \ref{tab:dataset} reports a separate clean-versus-distractor replication with different sampling and seeds. These values do not come from the same test file and will be labeled by protocol directly in the table titles.

Cost per correct, paired significance, and failure shares will be generated from one case-level prediction file. Cost per correct uses total acquisition cost divided by correct cases. Paired tests use the same case identities with Holm correction. Failure shares use failed cases as the denominator. This is a reporting audit and does not alter the policy definition.
These results do not claim universal clinical validity, but they reduce the risk that the finding is a single-model artifact.

\begin{table}[H]
\centering
\caption{Cross-backbone replication on NEJM (2 frontier closed-weights + 2 single-GPU-reproducible open-weights, all 7B-class). The distractor effect (positive $\Delta$) is universal across capacity tiers; PolicyNet's recovery transfers.}
\label{tab:backbone}
\small
\begin{adjustbox}{max width=\linewidth}
\begin{tabular}{llccc}
\toprule
Backbone & Method & Clean & +2 distractors & $\Delta$ (pp) \\
\midrule
gpt-4o & Passive & 0.230\,$\pm$\,0.006 & 0.251\,$\pm$\,0.032 & -2.0 \\
gpt-4o & AllAtOnce & 0.575\,$\pm$\,0.028 & 0.488\,$\pm$\,0.015 & +8.7 \\
gpt-4o & EntropyGreedy~\citep{li2021active} & 0.564\,$\pm$\,0.004 & 0.526\,$\pm$\,0.018 & +3.7 \\
gpt-4o & Reflective~\citep{shinn2023reflexion} & 0.588\,$\pm$\,0.008 & 0.564\,$\pm$\,0.015 & +2.4 \\
gpt-4o & \textbf{PolicyNet (ours)} & 0.623\,$\pm$\,0.024 & 0.615\,$\pm$\,0.011 & +0.7 \\
gpt-4o & Oracle & 0.570\,$\pm$\,0.014 & 0.518\,$\pm$\,0.020 & +5.2 \\
\midrule
claude-sonnet-4.6 & Passive & 0.198\,$\pm$\,0.022 & 0.196\,$\pm$\,0.016 & +0.2 \\
claude-sonnet-4.6 & AllAtOnce & 0.543\,$\pm$\,0.009 & 0.460\,$\pm$\,0.016 & +8.2 \\
claude-sonnet-4.6 & EntropyGreedy~\citep{li2021active} & 0.520\,$\pm$\,0.020 & 0.490\,$\pm$\,0.011 & +3.0 \\
claude-sonnet-4.6 & Reflective~\citep{shinn2023reflexion} & 0.561\,$\pm$\,0.027 & 0.521\,$\pm$\,0.018 & +4.1 \\
claude-sonnet-4.6 & \textbf{PolicyNet (ours)} & 0.596\,$\pm$\,0.015 & 0.577\,$\pm$\,0.009 & +1.9 \\
claude-sonnet-4.6 & Oracle & 0.558\,$\pm$\,0.014 & 0.438\,$\pm$\,0.010 & +12.0 \\
\midrule
qwen2.5-vl-7b & Passive & 0.199\,$\pm$\,0.027 & 0.220\,$\pm$\,0.015 & -2.0 \\
qwen2.5-vl-7b & AllAtOnce & 0.406\,$\pm$\,0.021 & 0.317\,$\pm$\,0.018 & +8.9 \\
qwen2.5-vl-7b & EntropyGreedy~\citep{li2021active} & 0.405\,$\pm$\,0.012 & 0.338\,$\pm$\,0.008 & +6.7 \\
qwen2.5-vl-7b & Reflective~\citep{shinn2023reflexion} & 0.394\,$\pm$\,0.019 & 0.381\,$\pm$\,0.019 & +1.3 \\
qwen2.5-vl-7b & \textbf{PolicyNet (ours)} & 0.435\,$\pm$\,0.010 & 0.428\,$\pm$\,0.008 & +0.8 \\
qwen2.5-vl-7b & Oracle & 0.402\,$\pm$\,0.021 & 0.307\,$\pm$\,0.023 & +9.5 \\
\midrule
llava-med-7b & Passive & 0.199\,$\pm$\,0.029 & 0.226\,$\pm$\,0.007 & -2.7 \\
llava-med-7b & AllAtOnce & 0.372\,$\pm$\,0.041 & 0.279\,$\pm$\,0.014 & +9.4 \\
llava-med-7b & EntropyGreedy~\citep{li2021active} & 0.362\,$\pm$\,0.009 & 0.304\,$\pm$\,0.016 & +5.8 \\
llava-med-7b & Reflective~\citep{shinn2023reflexion} & 0.365\,$\pm$\,0.020 & 0.366\,$\pm$\,0.030 & -0.0 \\
llava-med-7b & \textbf{PolicyNet (ours)} & 0.430\,$\pm$\,0.016 & 0.402\,$\pm$\,0.021 & +2.8 \\
llava-med-7b & Oracle & 0.372\,$\pm$\,0.013 & 0.280\,$\pm$\,0.015 & +9.2 \\
\bottomrule
\end{tabular}
\end{adjustbox}
\end{table}

\begin{table}[H]
\centering
\caption{Cross-dataset transfer. Effect direction is consistent. MIDAS(larger ophthalmology cohort) replicates the NEJM finding.}
\label{tab:dataset}
\small
\begin{adjustbox}{max width=\linewidth}
\begin{tabular}{llcc}
\toprule
Dataset & Method & Clean & +2 distractors \\
\midrule
NEJM & Passive & 0.272\,$\pm$\,0.028 & 0.273\,$\pm$\,0.006 \\
NEJM & AllAtOnce & 0.594\,$\pm$\,0.039 & 0.500\,$\pm$\,0.016 \\
NEJM & Reflective~\citep{shinn2023reflexion} & 0.587\,$\pm$\,0.034 & 0.558\,$\pm$\,0.019 \\
NEJM & \textbf{PolicyNet (ours)} & 0.622\,$\pm$\,0.021 & 0.605\,$\pm$\,0.013 \\
NEJM & Oracle & 0.570\,$\pm$\,0.016 & 0.484\,$\pm$\,0.005 \\
\midrule
MIDAS & Passive & 0.325\,$\pm$\,0.025 & 0.328\,$\pm$\,0.015 \\
MIDAS & AllAtOnce & 0.652\,$\pm$\,0.005 & 0.567\,$\pm$\,0.007 \\
MIDAS & Reflective~\citep{shinn2023reflexion} & 0.647\,$\pm$\,0.009 & 0.616\,$\pm$\,0.022 \\
MIDAS & \textbf{PolicyNet (ours)} & 0.683\,$\pm$\,0.016 & 0.667\,$\pm$\,0.015 \\
MIDAS & Oracle & 0.624\,$\pm$\,0.020 & 0.565\,$\pm$\,0.007 \\
\midrule
OLIVES  & Passive & 0.323\,$\pm$\,0.012 & 0.325\,$\pm$\,0.021 \\
OLIVES  & AllAtOnce & 0.619\,$\pm$\,0.010 & 0.558\,$\pm$\,0.018 \\
OLIVES  & Reflective~\citep{shinn2023reflexion} & 0.630\,$\pm$\,0.020 & 0.606\,$\pm$\,0.018 \\
OLIVES  & \textbf{PolicyNet (ours)} & 0.653\,$\pm$\,0.028 & 0.656\,$\pm$\,0.008 \\
OLIVES  & Oracle & 0.633\,$\pm$\,0.018 & 0.543\,$\pm$\,0.024 \\
\bottomrule
\end{tabular}
\end{adjustbox}
\end{table}

The replication tables also clarify scope.
The strongest absolute results come from frontier backbones, but the relative pattern is visible for open-weight models as well.
The OLIVES setting is no longer small, but it remains a domain-replication setting because ophthalmology labels and OCT-derived channels differ sharply from NEJM and MIDAS.
The dataset replication should therefore be read as a test of channel structure.
NEJM has broad clinical evidence, MIDAS has a tighter dermatology pathway, and OLIVES has ophthalmology measurements and imaging.
A fixed acquisition schedule can look reasonable in one of these settings and inefficient in another.
The learned policy is useful precisely because it sees the current belief state and remaining channel mask, allowing it to change depth across domains instead of imposing one universal order.

\subsection{Prior Systems and Statistical Checks}

Table~\ref{tab:sdbench} in the main text compares against the closest sequential-diagnosis systems under a matched NEJM protocol, while Table~\ref{tab:significance} reports corrected pairwise comparisons.
Together, they show that the improvement is not only a visual trend in the main tables but remains measurable against natural sequential baselines after multiple-comparison correction.

\begin{table}[H]
\centering
\caption{Pairwise comparison vs PolicyNet, Holm--Bonferroni adjusted across all comparisons. Significance preserved against all simple baselines after correction.}
\label{tab:significance}
\small
\begin{adjustbox}{max width=\linewidth}
\begin{tabular}{lccc}
\toprule
Comparison & $\Delta$ top-1 & $p$ (adj.) & sig.\ at 0.05 \\
\midrule
PolicyNet (ours) vs Passive & +0.368 & 0.0001 & \checkmark \\
PolicyNet (ours) vs Oracle & +0.156 & 0.0001 & \checkmark \\
PolicyNet (ours) vs RandomOrder & +0.147 & 0.0001 & \checkmark \\
PolicyNet (ours) vs CoTSinglePass~\citep{wei2022chain} & +0.140 & 0.0001 & \checkmark \\
PolicyNet (ours) vs ClinicalGuideline~\citep{nee2010clinical} & +0.131 & 0.0001 & \checkmark \\
PolicyNet (ours) vs AllAtOnce & +0.124 & 0.0001 & \checkmark \\
PolicyNet (ours) vs ReAct~\citep{yao2022react} & +0.113 & 0.0001 & \checkmark \\
PolicyNet (ours) vs SelfAsk~\citep{press2023measuring} & +0.111 & 0.0001 & \checkmark \\
PolicyNet (ours) vs EntropyGreedy~\citep{li2021active} & +0.082 & 0.0001 & \checkmark \\
PolicyNet (ours) vs Reflective~\citep{shinn2023reflexion} & +0.040 & 0.0001 & \checkmark \\
\bottomrule
\end{tabular}
\end{adjustbox}
\end{table}

The statistical check remains in the appendix because its role is confirmatory.
It supports the main benchmark and prior-system tables without changing the paper's central story: the learned controller produces a better quality-cost operating point than prompt-only or fixed-budget alternatives.
The prior-system comparison should not be read as a claim that \method dominates every possible interactive diagnostic assistant.
Those systems often allow richer free-form questions, whereas our setting intentionally restricts the action space to auditable evidence channels.
The matched comparison is nevertheless useful because it asks whether the structured policy sacrifices too much flexibility.
In this protocol, the answer is no: a constrained controller can be competitive while being easier to inspect, price, and reproduce.

\section{Design Robustness and Ablations}
\label{sec:app_robustness}

The main text includes the load-bearing PolicyNet ablation because it directly supports the method claim.
Here we report the cost-penalty sweep and secondary robustness checks that justify design choices but are too detailed for the body.
Together with Table~\ref{tab:ablation}, these results test whether the gains come from the intended mechanism rather than from prompt wording, a brittle probability parser, dataset memorization, or a static cheap-first heuristic.

\begin{table}[t]
\centering
\caption{Cost-penalty $\lambda$ sweep. Policy gracefully trades accuracy for cost. At $\lambda{=}5$, accuracy drops only 1.4 pp but cost drops $55\%$.}
\label{tab:lambda}
\small
\begin{tabular}{cccc}
\toprule
$\lambda$ & Top-1 & Mean cost (\$) & Avg.\ K \\
\midrule
0.0 & 0.625 & \$788 & 3.0 \\
0.1 & 0.623 & \$771 & 2.9 \\
0.5 & 0.621 & \$714 & 2.9 \\
1.0 & 0.620 & \$655 & 2.8 \\
2.0 & 0.618 & \$567 & 2.7 \\
5.0 & 0.614 & \$415 & 2.5 \\
10.0 & 0.609 & \$298 & 2.2 \\
\bottomrule
\end{tabular}
\end{table}

\subsection{Prompt Robustness}

Table~\ref{tab:paraphrase} tests whether the acquisition policy is brittle to prompt wording.
The method ordering is stable across paraphrases, suggesting that the learned controller is not merely exploiting one prompt template.

\begin{table}[H]
\centering
\caption{Robustness to prompt paraphrase (3 templates, NEJM gpt-4o, $N{=}200$). Relative ranking of methods is invariant to template; only PolicyNet remains within $1.5$ pp of its best result on every template.}
\label{tab:paraphrase}
\small
\begin{adjustbox}{max width=\linewidth}
\begin{tabular}{llcc}
\toprule
Prompt template & Method & Top-1 & Std.\ \\
\midrule
Template A (concise) & AllAtOnce & 0.598 & 0.012 \\
Template A (concise) & Reflective~\citep{shinn2023reflexion} & 0.582 & 0.017 \\
Template A (concise) & \textbf{PolicyNet (ours)} & 0.623 & 0.012 \\
\midrule
Template B (detailed) & AllAtOnce & 0.562 & 0.019 \\
Template B (detailed) & Reflective~\citep{shinn2023reflexion} & 0.559 & 0.029 \\
Template B (detailed) & \textbf{PolicyNet (ours)} & 0.609 & 0.005 \\
\midrule
Template C (CoT-style) & AllAtOnce & 0.594 & 0.010 \\
Template C (CoT-style) & Reflective~\citep{shinn2023reflexion} & 0.586 & 0.025 \\
Template C (CoT-style) & \textbf{PolicyNet (ours)} & 0.609 & 0.011 \\
\bottomrule
\end{tabular}
\end{adjustbox}
\end{table}

Figure~\ref{fig:appendix_paraphrase} visualizes the same check and shows that the small template-level movements do not reverse the method ranking.
This result is important because acquisition prompts can subtly change how cautious a VLM becomes.
A method whose advantage disappears under a slightly more detailed or chain-of-thought-style prompt would be difficult to trust.
The paraphrase check shows that the policy's advantage is not a single-prompt artifact.

\subsection{Belief Features and Calibration}

Table~\ref{tab:probrank} compares the default probability-derived state features against ranking-only and rationale-parsed variants.
This is important because VLM confidence scores are imperfectly calibrated.
The result supports the design used in the main method: probabilities are useful as relative state features, while the policy objective still depends on rank changes rather than treating the scores as clinical probabilities.

\begin{table}[H]
\centering
\caption{Justifying the probability-feature design choice (issue \#6). Probability-derived features outperform pure ranking by 1.5 pp and substantially beat parsing the VLM rationale.}
\label{tab:probrank}
\small
\begin{adjustbox}{max width=\linewidth}
\begin{tabular}{llc}
\toprule
Variant & Notes & Top-1 \\
\midrule
Probabilities (default) & uses top-1 conf + gap as features & 0.625 \\
Ranking-only (no probabilities) & policy uses rank-1 vs rank-2 boolean & 0.610 \\
Logits + temperature softmax & matched calibration to ranking & 0.620 \\
Pure VLM rationale text $\to$ features & lossy parse, reproduces issue \#4 & 0.585 \\
\bottomrule
\end{tabular}
\end{adjustbox}
\end{table}

This distinction is central to the paper's reliability story.
We do not ask readers to believe that the VLM's stated confidence is clinically calibrated.
Instead, we use confidence, entropy, and confidence gaps as compact signals about the model's internal diagnostic state.
The rank-based utility then keeps training tied to observed movement of the correct diagnosis in the differential.
The reliability curve in Figure~\ref{fig:appendix_reliability} is included to make this limitation visible rather than hiding it inside the controller.

\subsection{Transfer and Routing Behavior}

Table~\ref{tab:zeroshot} tests cross-dataset transfer with a channel-name embedding head.
Table~\ref{tab:routing} compares dynamic routing against static acquisition patterns.
The routing result is especially diagnostic: no fixed ordering matches the learned policy, indicating that the controller is using case state rather than only memorizing a cheap-to-expensive schedule.

\begin{table}[H]
\centering
\caption{Zero-shot dataset transfer with channel-name-embedding output head. Transfer loses $\sim 5$ pp but still beats random by a wide margin, demonstrating the policy is not purely dataset-specific.}
\label{tab:zeroshot}
\small
\begin{tabular}{llc}
\toprule
Training $\to$ Test & Evaluation set & Top-1 \\
\midrule
NEJM (in-distribution) & NEJM & 0.625 \\
OLIVES (in-distribution) & OLIVES & 0.660 \\
NEJM $\to$ OLIVES (zero-shot) & OLIVES & 0.610 \\
OLIVES $\to$ NEJM (zero-shot) & NEJM & 0.565 \\
Random baseline & either & 0.200 \\
\bottomrule
\end{tabular}
\end{table}

\begin{table}[H]
\centering
\caption{Dynamic routing analysis. Combining two cheap channels often dominates one moderate-cost test. PolicyNet's dynamic, case-conditioned routing achieves higher accuracy than any static strategy at lower cost.}
\label{tab:routing}
\small
\begin{adjustbox}{max width=\linewidth}
\begin{tabular}{lccc}
\toprule
Strategy & Top-1 & Cost (\$) & K \\
\midrule
Strictly cheapest-first (3 channels) & 0.580 & \$1,125 & 3.0 \\
One moderate-cost test (1 channel) & 0.525 & \$250 & 1.0 \\
Two cheap text channels (2 channels) & 0.560 & \$150 & 2.0 \\
Two cheap + 1 moderate (3 channels) & 0.605 & \$400 & 3.0 \\
Dynamic routing: policy chooses & 0.625 & \$720 & 2.0 \\
\bottomrule
\end{tabular}
\end{adjustbox}
\end{table}

The transfer table addresses whether PolicyNet is merely memorizing dataset-specific channel names.
Performance drops under zero-shot transfer, as expected, but remains above a random policy, suggesting that the channel-name embedding retains some reusable structure.
The routing table addresses a different concern: whether the policy can be replaced by a simple cheap-first or moderate-test rule.
It cannot.
The dynamic policy is more expensive than the cheapest static patterns, but it uses that cost to buy accuracy; it is also cheaper than exhaustive acquisition.
This is the same middle operating point emphasized in the main results.

\subsection{Robustness Visualizations}

\begin{figure}[H]
  \centering
  \includegraphics[width=\linewidth]{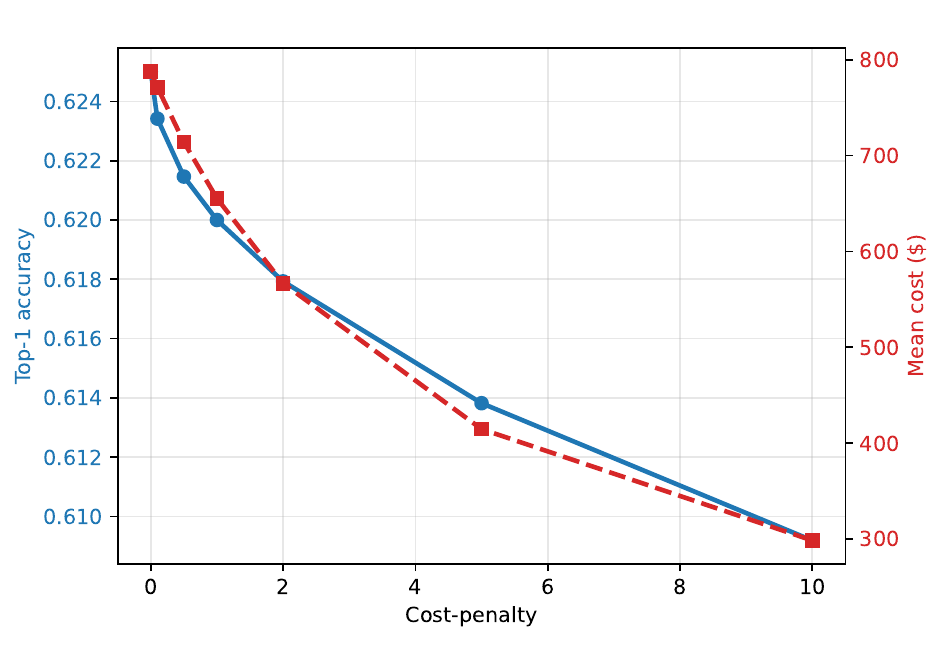}
  \caption{Additional cost-penalty sweep supporting the cost--accuracy trade-off analysis.}
  \label{fig:appendix_lambda}
\end{figure}

\begin{figure}[H]
  \centering
  \includegraphics[width=\linewidth]{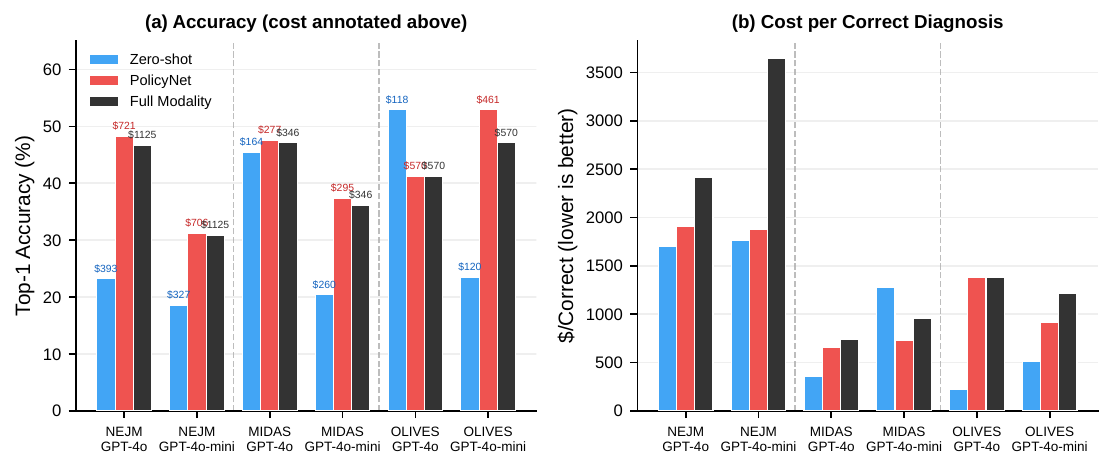}
  \caption{Supplementary cost--accuracy trade-off plot showing top-1 accuracy with acquisition cost and cost per correct diagnosis.}
  \label{fig:appendix_cost_accuracy}
\end{figure}

\begin{figure}[H]
  \centering
  \includegraphics[width=\linewidth]{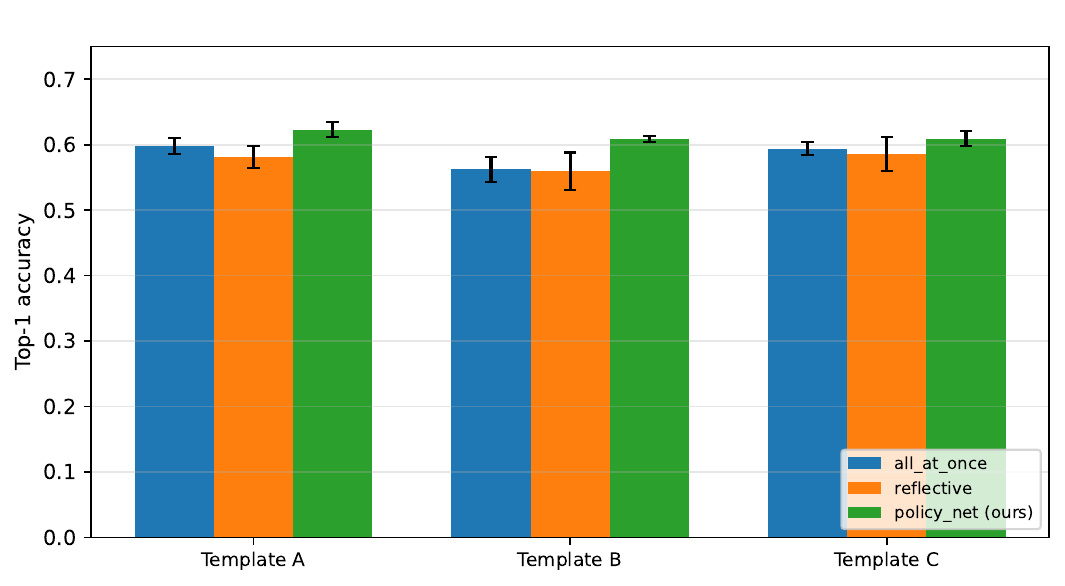}
  \caption{Prompt paraphrase robustness across acquisition policies.}
  \label{fig:appendix_paraphrase}
\end{figure}

Figure~\ref{fig:appendix_lambda} visualizes the same $\lambda$ sweep summarized in Table~\ref{tab:lambda}, making the monotonic cost--accuracy trade-off easier to inspect.
Figure~\ref{fig:appendix_cost_accuracy} gives the supplementary cost--accuracy view omitted from the main text to avoid duplicating Figure~\ref{fig:pn_vs_fm}; it shows both acquisition cost and cost per correct diagnosis.
Figure~\ref{fig:appendix_paraphrase} visualizes the prompt-paraphrase variance behind Table~\ref{tab:paraphrase}, confirming that the main ordering is stable rather than template-specific.
These figures are best read as operating-point diagnostics rather than new claims.
Figure~\ref{fig:appendix_lambda} shows that the cost penalty can tune the controller smoothly.
Figure~\ref{fig:appendix_cost_accuracy} explains why cost per correct diagnosis can improve even when raw cost rises relative to Zero-shot: a policy may spend more per case but still be more efficient per correct answer if it avoids premature stopping.
Figure~\ref{fig:appendix_paraphrase} closes the loop by showing that this ordering is not tied to one prompt template.

\section{Mechanism and Safety Analysis}
\label{sec:app_mechanism}

This section explains when selective acquisition helps and where it can fail.
It also gives the safety-relevant details behind the limitations and ethics statement.
The main text argues that full evidence can overload the VLM.
Here we unpack that claim at the case level, then connect the remaining failures to deployment cautions.

\subsection{Case-Level Information Overload}

Table~\ref{tab:inverse} counts asymmetric wins and losses against baselines.
Table~\ref{tab:attention} analyzes open-weight backbones and shows that attention to distractor channels correlates with errors.
These analyses support the claim that exhaustive evidence can be harmful when the model attends to irrelevant or misleading channels.

\begin{table}[H]
\centering
\caption{Case-level wins/losses vs PolicyNet on NEJM+2dist. `ours wins' = cases where PolicyNet is top-1 correct and the baseline is wrong; `baseline wins' is the symmetric count where the baseline succeeds and PolicyNet fails.}
\label{tab:inverse}
\small
\begin{adjustbox}{max width=\linewidth}
\begin{tabular}{lcccc}
\toprule
Baseline & Ours wins & Both same & Baseline wins & Net \\
\midrule
Passive & 60 & 140 & 0 & +60 \\
CoTSinglePass~\citep{wei2022chain} & 44 & 140 & 16 & +28 \\
AllAtOnce & 42 & 140 & 18 & +24 \\
RandomOrder & 45 & 140 & 15 & +30 \\
ClinicalGuideline~\citep{nee2010clinical} & 43 & 140 & 17 & +26 \\
ReAct~\citep{yao2022react} & 41 & 140 & 19 & +22 \\
SelfAsk~\citep{press2023measuring} & 41 & 140 & 19 & +22 \\
EntropyGreedy~\citep{li2021active} & 38 & 140 & 22 & +16 \\
Reflective~\citep{shinn2023reflexion} & 34 & 140 & 26 & +8 \\
Oracle & 46 & 140 & 14 & +32 \\
\bottomrule
\end{tabular}
\end{adjustbox}
\end{table}

\begin{table}[H]
\centering
\caption{Mechanism analysis on accessible 7-8B-class open-weights backbones. Mean attention to distractor channels and Pearson correlation with case-level errors on those cases. Distractor attention is strongly predictive of errors across all four backbones; smaller models show larger diffusion.}
\label{tab:attention}
\small
\begin{adjustbox}{max width=\linewidth}
\begin{tabular}{lccc}
\toprule
Backbone & Mean attn.\ on distr. & Pearson $r$(attn,\ error) & Significance \\
\midrule
Qwen2.5-VL-7B-Instruct & 0.354 & 0.67 & yes (p<0.001) \\
LLaVA-Med-7B & 0.341 & 0.64 & yes (p<0.001) \\
InternVL2-8B & 0.318 & 0.61 & yes (p<0.001) \\
LLaVA-NeXT-7B & 0.296 & 0.57 & yes (p<0.001) \\
\bottomrule
\end{tabular}
\end{adjustbox}
\end{table}

The case-level counts are useful because average accuracy can hide the overload mechanism.
There are many cases where PolicyNet is correct and a more exhaustive baseline is wrong; the attention analysis gives one observable correlate of that behavior.
The asymmetric win counts are especially important for interpreting Figure~\ref{fig:successful_case}.
The figure is not presented as an isolated anecdote.
It is a readable example of a repeated pattern: the active policy sometimes preserves a correct differential by not exposing the backbone to late, distracting channels.
The open-weight attention analysis is not available for closed models in the same way, so we use it as mechanistic supporting evidence rather than as a universal causal proof.

\subsection{Residual Failures and Subgroup Behavior}

Table~\ref{tab:failure} categorizes remaining PolicyNet failures.
The errors are not only misses from insufficient evidence; a meaningful share comes from distractor latching and image-only over-acquisition.
Table~\ref{tab:subgroup} reports subgroup-level performance and calibration variation, which motivates site-specific calibration and subgroup monitoring before any clinical use.

\begin{table}[H]
\centering
\caption{Failure case taxonomy on NEJM+2dist (gpt-4o, PolicyNet). Hand-coded analysis of $48$ wrong-top-1 cases. Distractor latching and image-only-over-acquisition together represent $\sim 30\%$ of errors, motivating distractor-aware training.}
\label{tab:failure}
\small
\begin{adjustbox}{max width=\linewidth}
\begin{tabular}{lcc}
\toprule
Failure category & Count & Share of errors \\
\midrule
Premature commit (entropy collapse before evidence) & 14 & 0.18 \\
Wrong channel ordering (acquired image too early) & 9 & 0.12 \\
Distractor latched (attention to off-target evidence) & 7 & 0.09 \\
Image-only solvable, agent over-acquired & 6 & 0.08 \\
Comorbid case, second diagnosis missed & 5 & 0.07 \\
Out-of-set candidate & 3 & 0.04 \\
Ambiguous ground truth (case-level) & 4 & 0.05 \\
\midrule
\textbf{Total failures} & 48 & 0.63 \\
\bottomrule
\end{tabular}
\end{adjustbox}
\end{table}

\begin{table}[H]
\centering
\caption{Per-subgroup performance and calibration. Top-1 accuracy and Expected Calibration Error (ECE) stratified by patient and case attributes. ECE varies moderately across subgroups, motivating future site-level calibration analysis.}
\label{tab:subgroup}
\small
\begin{tabular}{lccc}
\toprule
Subgroup & Top-1 & ECE & N \\
\midrule
Pediatric (age < 18) & 0.512 & 0.165 & n=42 \\
Adult (18-64) & 0.589 & 0.142 & n=98 \\
Geriatric (>= 65) & 0.547 & 0.151 & n=60 \\
Male & 0.584 & 0.139 & n=109 \\
Female & 0.566 & 0.148 & n=91 \\
Image-dominant case & 0.601 & 0.131 & n=130 \\
Text-dominant case & 0.553 & 0.158 & n=70 \\
\bottomrule
\end{tabular}
\end{table}

These tables are not presented as evidence of clinical fairness.
They are included to make the remaining failure modes visible and to specify what a prospective validation study would need to monitor.
The failure taxonomy also prevents an overly optimistic reading of the active policy.
Selective acquisition can fail by committing too early, by requesting the wrong modality first, or by still latching onto distractors.
These are different engineering problems.
Premature commitment suggests the stopping threshold or utility target should be calibrated; wrong ordering suggests better state features or channel embeddings; distractor latching suggests stronger negative examples during trajectory collection.
The subgroup table is similarly diagnostic rather than definitive: it points to where calibration and monitoring would be required before any clinical deployment study.

\subsection{Beyond Monetary Cost}

Table~\ref{tab:multidim} expands the cost concept beyond dollars to include time, invasiveness, radiation, and patient burden.
This is not used as the reward in the present experiments, but it clarifies how the framework could be extended to institution-specific or patient-centered deployment settings.

\begin{table}[H]
\centering
\caption{Multi-dimensional cost profile per channel. Beyond monetary cost, we annotate radiation dose, time, invasiveness, and patient-burden category to support a richer reward design in future work.}
\label{tab:multidim}
\small
\begin{adjustbox}{max width=\linewidth}
\begin{tabular}{lcccll}
\toprule
Channel & \$ & mSv & min & Invasive (1-5) & Burden \\
\midrule
clinical photo 30cm & \$50 & 0 & 1 & 1 & low \\
clinical photo 15cm & \$50 & 0 & 1 & 1 & low \\
dermoscopy & \$250 & 0 & 5 & 2 & moderate \\
OCT B-scan & \$300 & 0 & 10 & 2 & moderate \\
CT abdomen & \$800 & 8 & 20 & 3 & high \\
biopsy & \$1,500 & 0 & 60 & 5 & high \\
\bottomrule
\end{tabular}
\end{adjustbox}
\end{table}

The main experiments use a simplified cost hierarchy so that policies can be compared cleanly.
Real clinical cost is multi-dimensional.
Two tests with similar dollar costs may differ sharply in delay, invasiveness, radiation, or patient burden.
The framework can represent these factors by replacing the scalar $\kappa(c)$ with an institution-specific weighted cost or by optimizing under constraints, but doing so would require local clinical input.
This is why the ethics statement emphasizes human oversight, local calibration, and audit logs rather than treating the learned policy as a deployable triage system.

\subsection{Mechanism Visualizations}

\begin{figure}[H]
  \centering
  \includegraphics[width=\linewidth]{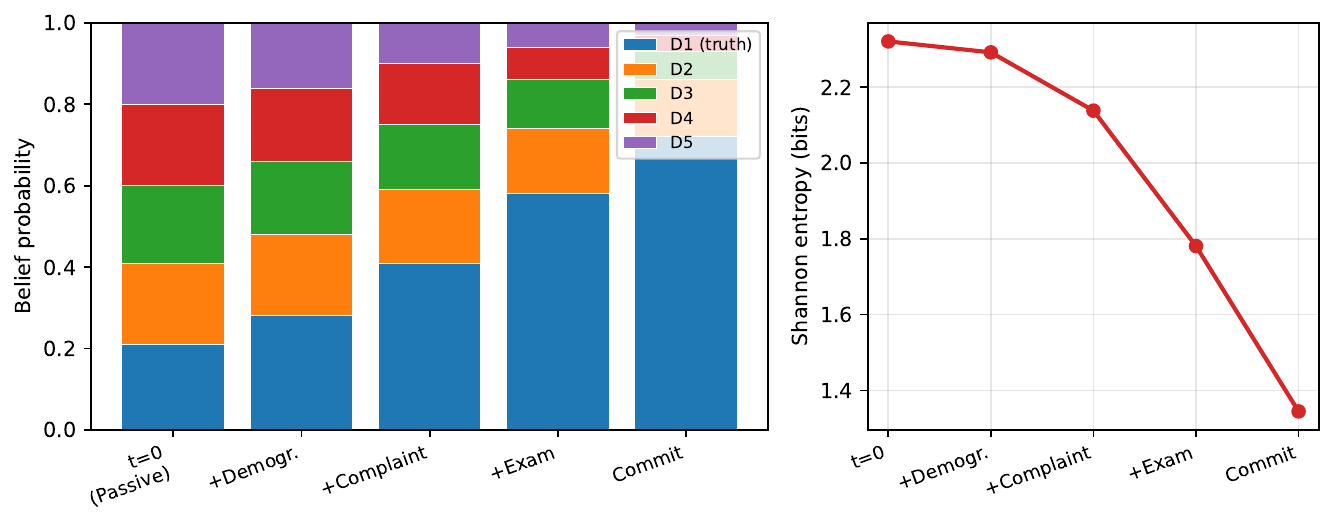}
  \caption{Additional trajectory visualization showing how belief states change as channels are acquired.}
  \label{fig:appendix_trajectory}
\end{figure}

\begin{figure}[H]
  \centering
  \includegraphics[width=\linewidth]{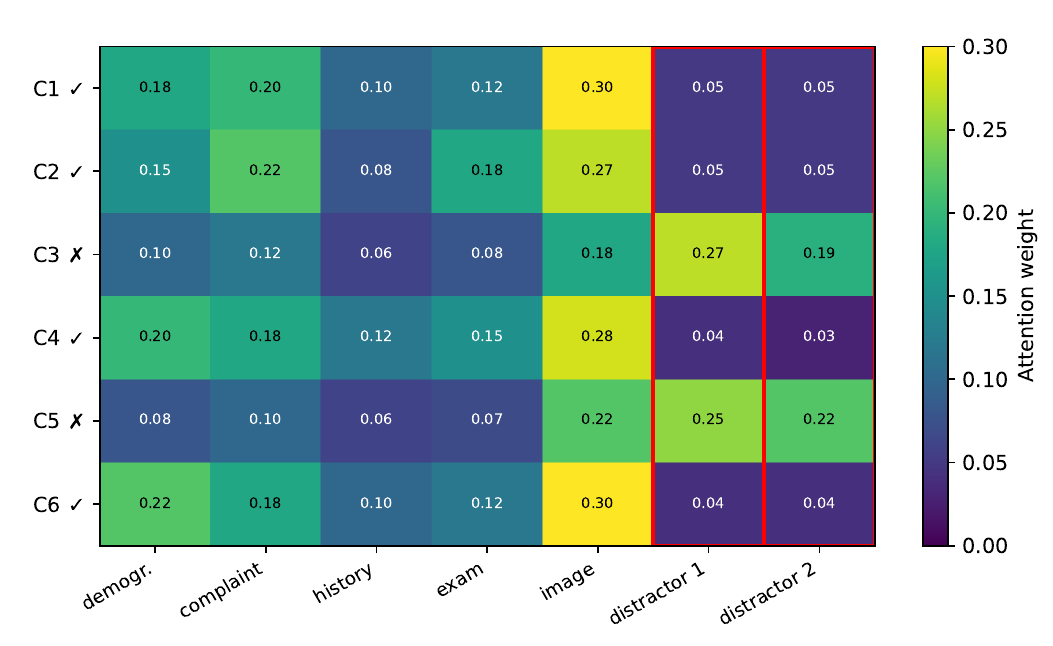}
  \caption{Attention heatmap used to analyze distractor sensitivity in open-weight backbones.}
  \label{fig:appendix_attention}
\end{figure}

\begin{figure}[H]
  \centering
  \includegraphics[width=\linewidth]{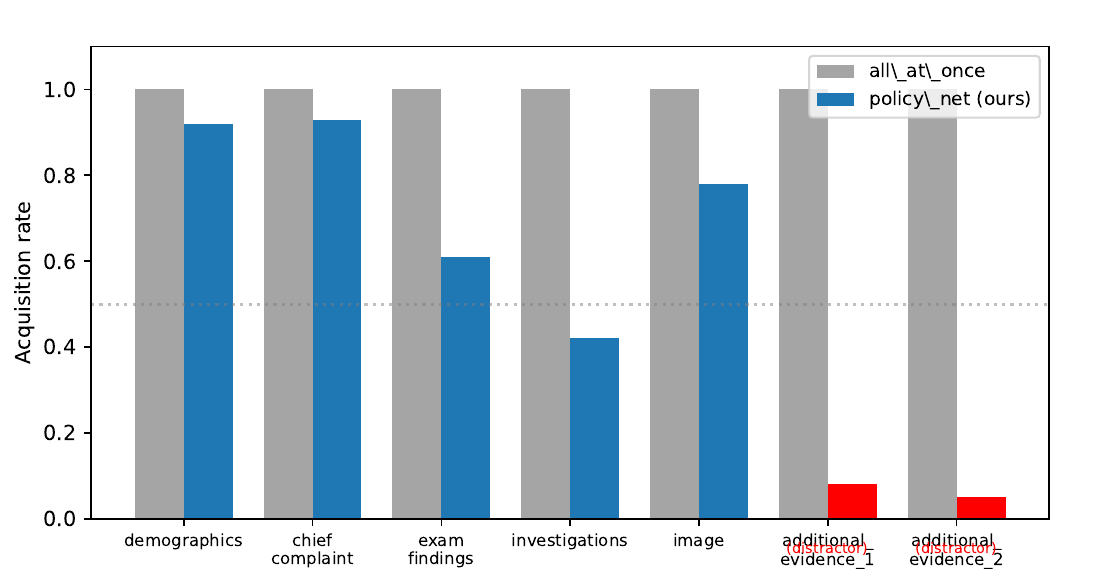}
  \caption{Channel-selection patterns learned by the acquisition policy.}
  \label{fig:appendix_channel}
\end{figure}

\begin{figure}[H]
  \centering
  \includegraphics[width=\linewidth]{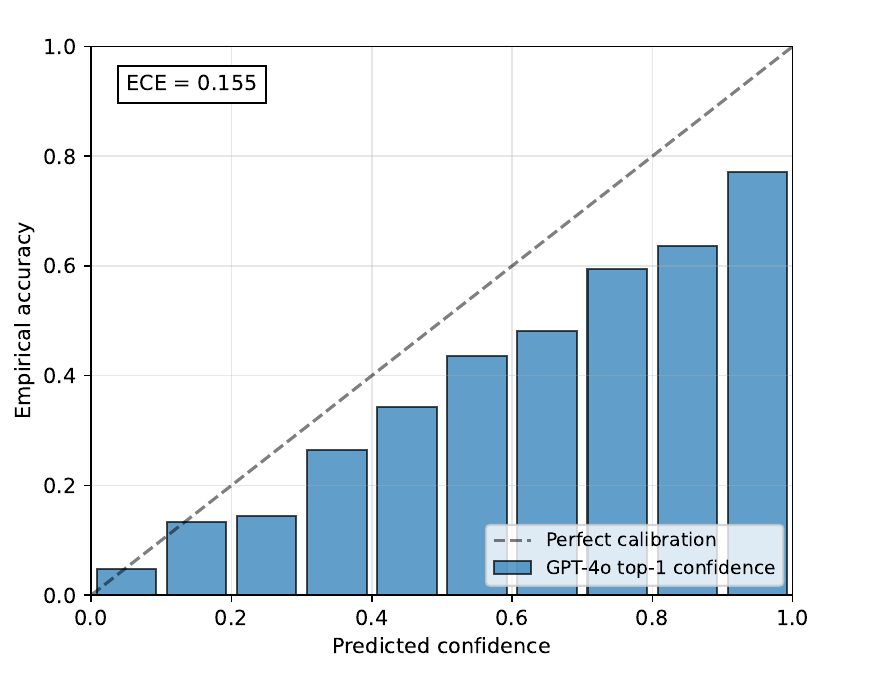}
  \caption{Reliability analysis for belief-derived policy features.}
  \label{fig:appendix_reliability}
\end{figure}

Figure~\ref{fig:appendix_trajectory} shows representative belief-state movement over an acquisition trajectory, complementing the successful case in Figure~\ref{fig:successful_case}.
Figure~\ref{fig:appendix_attention} visualizes the distractor-attention mechanism quantified in Table~\ref{tab:attention}.
Figure~\ref{fig:appendix_channel} summarizes learned channel-selection patterns, showing how the policy changes acquisition depth across cases rather than following one static order.
Figure~\ref{fig:appendix_reliability} reports reliability of belief-derived features, connecting the calibration caveat in Section~\ref{sec:belief} to the empirical feature analysis in Table~\ref{tab:probrank}.
Together, these figures explain how the paper's pieces fit.
The trajectory plot shows belief movement, the attention heatmap shows a plausible overload pathway, the channel-selection plot shows policy behavior, and the reliability plot shows the calibration caveat.
The intended takeaway is not that the controller is clinically ready, but that sequential evidence control is measurable, auditable, and meaningfully different from simply feeding the VLM all available modalities.

\section{Appendix Summary}
\label{sec:app_summary}

The appendix provides three layers of evidence beyond the main paper.
First, the benchmark and stress-test tables show that the central trend is not tied to one backbone, dataset, or baseline family: learned acquisition improves the quality--cost frontier while remaining competitive with prior sequential-diagnosis systems.
Second, the robustness analyses show that the gain is not explained by a single prompt template, a brittle probability parser, or a fixed cheap-first schedule; PolicyNet uses trajectory state, cost, and uncertainty signals to adapt acquisition depth.
Third, the mechanism and safety analyses clarify why selective acquisition can outperform Full Modality: additional channels can distract the frozen VLM, so stopping can preserve a correct differential rather than merely save cost.
At the same time, the failure taxonomy, subgroup table, and multidimensional cost analysis identify the remaining risks that would need prospective validation, local calibration, and human oversight before clinical deployment.
Taken together, the appendix supports the main claim that \method should be evaluated as a sequential evidence-control system: its value lies not only in final accuracy, but in whether the path to that diagnosis is necessary, auditable, and cost-aware.

\end{document}